\pdfoutput=1

\documentclass[11pt]{article}

\usepackage[]{acl}

\usepackage{times}
\usepackage{latexsym}
\usepackage{amsmath}
\usepackage{stmaryrd}
\usepackage{amsfonts}
\usepackage{multirow}
\usepackage{booktabs}
\usepackage{adjustbox}
\usepackage{array}
\usepackage{graphicx} 
\usepackage{xcolor}
\usepackage{makecell}
\usepackage{arydshln}
\usepackage{fdsymbol}
\definecolor{cyan}{rgb}{0.32, 0.64, 0.80}
\definecolor{orange}{rgb}{0.93, 0.46, 0.14}
\definecolor{green}{rgb}{0.49, 0.70, 0.35}
\definecolor{blue}{rgb}{0.22, 0.44, 0.80}
\definecolor{lightblue}{rgb}{0.12, 0.44, 0.80}
\definecolor{red}{rgb}{0.99, 0.02, 0.02}
\definecolor{lightred}{rgb}{0.65, 0.12, 0.12}
\definecolor{purple}{rgb}{0.60, 0.41, 0.61}
\definecolor{train}{rgb}{0.98, 0.92, 0.88}
\definecolor{test}{rgb}{0.88, 0.96, 1.0}

\usepackage[T1]{fontenc}

\usepackage[utf8]{inputenc}

\usepackage{microtype}

\usepackage{inconsolata}

\title{Why So Gullible? Enhancing the Robustness of\\ Retrieval-Augmented Models against Counterfactual Noise}

\author{Giwon Hong$^*$$^1$ \qquad Jeonghwan Kim$^*$$^2$ \qquad Junmo Kang$^*$$^3$ \\ \qquad \textbf{Sung-Hyon Myaeng}$^4$ \qquad \textbf{Joyce Jiyoung Whang}$^\dagger$$^4$ \\
  $^1$University of Edinburgh \qquad $^2$UIUC \qquad $^3$Georgia Tech \qquad $^4$KAIST \\
  \small{\texttt{\textmd{giwon.hong@ed.ac.uk}} \qquad \texttt{\textmd{jk100@illinois.edu}} \qquad \texttt{\textmd{junmo.kang@gatech.edu}}} \\
  \small{\texttt{\textmd{\{myaeng, jjwhang\}@kaist.ac.kr}}} \\
  \small{\textit{*Equal contribution}} \qquad
  \small{\textit{\dagger Corresponding author}}
}


\begin{document}
\maketitle
\begin{abstract}
Most existing retrieval-augmented language models (LMs) assume a na\"ive dichotomy within a retrieved document set: query-relevance and irrelevance. Our work investigates a more challenging scenario in which even the "relevant" documents may contain misleading or incorrect information, causing conflict among the retrieved documents and thereby negatively influencing model decisions as noise. We observe that existing LMs are highly brittle to the presence of conflicting information in both the fine-tuning and in-context few-shot learning scenarios. We propose approaches for handling knowledge conflicts among retrieved documents by explicitly fine-tuning a discriminator or prompting \mbox{GPT-3.5} to elicit its discriminative capability. Our empirical results on open-domain QA show that these approaches significantly enhance model robustness. We also provide our findings on incorporating the fine-tuned discriminator's decision into the in-context learning process, proposing a way to exploit the benefits of two disparate learning schemes. Alongside our findings, we provide \textsc{MacNoise}, a machine-generated, conflict-induced dataset to further encourage research in this direction\footnote{We release our code and dataset at: \url{https://github.com/wjdghks950/Discern-and-Answer}}.
\end{abstract}

\section{Introduction}
The general framework of retrieval-augmented language models (LMs) for question answering (QA) consists of retrieving documents related to a question using a sparse \cite{robertson2009probabilistic, jang-etal-2021-ultra} or a dense \cite{karpukhin2020dense} retriever, and processing the retrieved documents using encoder \cite{devlin2019bert} or decoder \cite{2020t5} models to derive an answer. Despite being used in many practical applications, most retrieval-augmented LMs \cite{pmlr-v119-guu20a, lewis2020retrieval, izacard2021leveraging, lewis-etal-2021-paq} are predicated on a na\"ive assumption: the retrieved documents are either relevant or irrelevant to the query. However, such a dichotomous view overlooks the fact that in real-world scenarios, the documents purportedly relevant to the query may not consistently offer accurate or reliable information, leading to conflicts among the retrieved documents. Such conflicts, as noise, can adversely affect the models that heavily rely on the veracity of the provided information.
Inconsistencies caused by conflicting information may occur for various reasons such as updated/outdated or fabricated/hallucinated information, with the latter being a significantly growing concern due to documents generated by large language models (LLMs) flooding the Web.

\begin{figure}[t]
    \centering
    \includegraphics[scale=0.31]{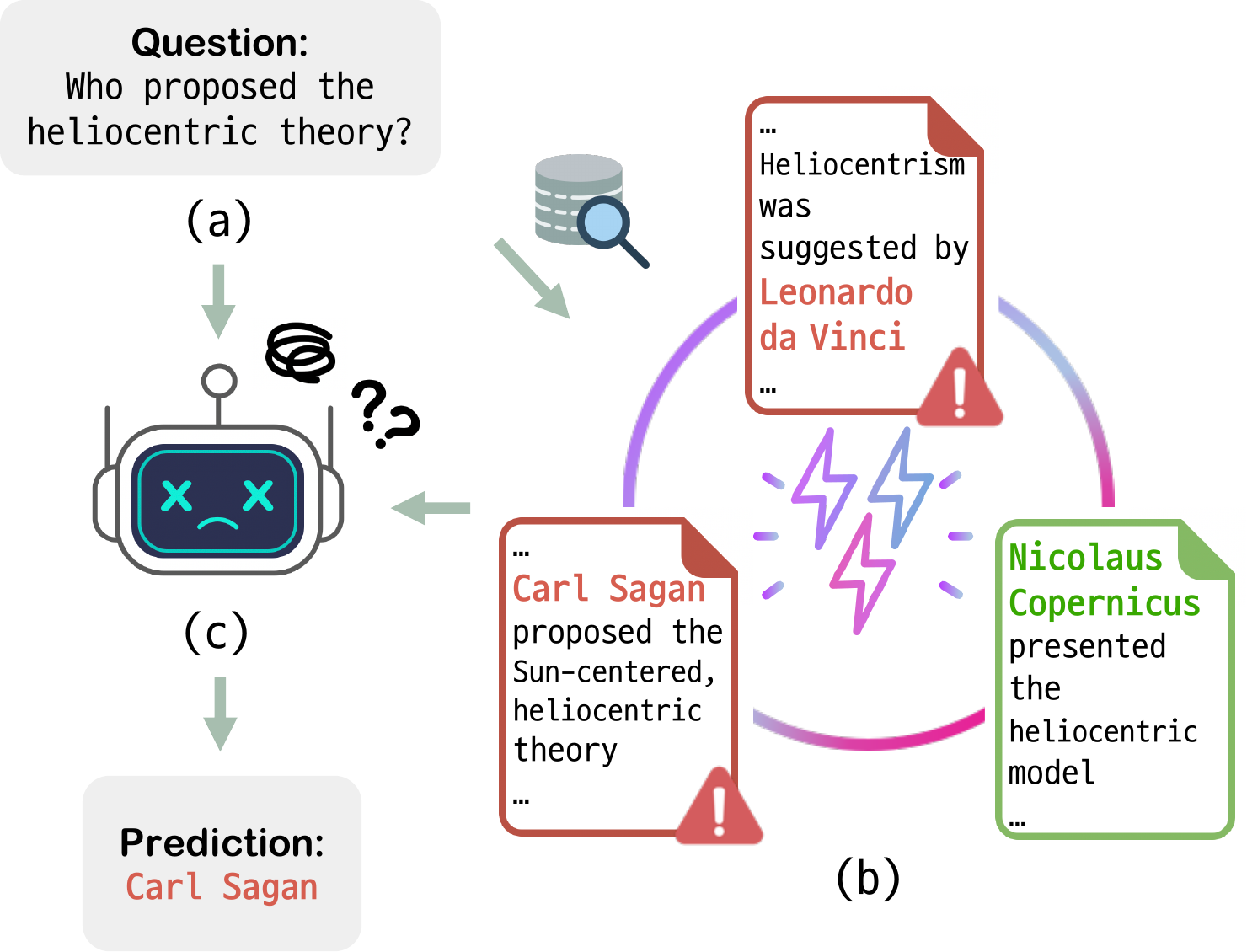}
    \caption{In an ODQA setting, \texttt{(a)} a question is used to retrieve a set of \texttt{(b)} relevant documents which may contain conflict-causing documents that render \texttt{(c)} the retrieval-augmented LMs unreliable.
    }
    \vspace{-0.2cm}
    \label{fig:figure1}
\end{figure}

We study the robustness of retrieval-augmented LMs in the presence of noise and the ensuing knowledge conflict in open-domain question answering (ODQA). 
To facilitate a controllable study, we adopt the widely-used \citet{longpre-etal-2021-entity}'s framework that deliberately perturbs the retrieved documents, which is also used in previous works on knowledge conflict \cite{chen2022rich, neeman-etal-2023-disentqa}.
This deliberate perturbation causes conflict among the documents, as shown in Figure \ref{fig:figure1}, which undermines the model's reliability even in the presence of a gold document.

Our empirical results show existing models such as FiD \cite{izacard2021leveraging} and GPT-3.5 (\texttt{text-davinci-003}) \cite{NEURIPS2020_1457c0d6} are highly susceptible to conflicting information. To alleviate this problem, we propose inducing the discrimination capabilities and exploiting them in the fine-tuned (FiD; \S \ref{sec:disc_train}) and in-context learned (GPT-3.5; \S \ref{prompting}) models to let them focus on reliable information. 
We demonstrate that (i) the fine-tuned LM achieves high precision in discerning authentic from counterfactual documents, and (ii) large language models (LLMs) leverage rich parametric knowledge to perform tasks with limited training data, but exhibit weakness in distinguishing noisy documents (\S \ref{sec:eval_longpre}).
Based on our findings, we combine the strengths of fine-tuning and prompting, highlighting the potential benefits of leveraging lightweight fine-tuned LMs to assist LLMs.

Furthermore, while previous works \cite{chen2022rich, neeman-etal-2023-disentqa, si2023prompting} also leverage \citet{longpre-etal-2021-entity} to emulate knowledge conflict scenarios, the simple entity-swap technique faces several limitations regarding the verisimilitude of the perturbed texts. To this end, we also release a set of LLM-generated contradictory documents using 
GPT-4 \cite{openai2023gpt4} to enable a more realistic and challenging study (\S \ref{sec:eval_gpt4}). We hope this can further encourage future works to explore conflict resolution in the retrieval-augmented LMs.
Our contributions include:
\begin{itemize}
    \vspace{-0.1cm}
    \itemsep-0.1em
    \item We highlight the vulnerability of retrieval-augmented models to counterfactual noise, irrespective of whether they are fine-tuned or in-context learned models.
    \item We propose a simple yet effective approach for enhancing discrimination capabilities so as to mitigate the model's susceptibility to noise.
    \item We construct a new LLM-generated counterfactual dataset, \textsc{MacNoise}, which turns out to be a challenging knowledge-conflict benchmark, as shown in our evaluation.
    \item Our work opens up a new direction for future works to integrate the benefits of both fine-tuning and in-context learning paradigms.
\end{itemize}

\begin{figure*}[th!]
    \centering
    \includegraphics[scale=0.43]{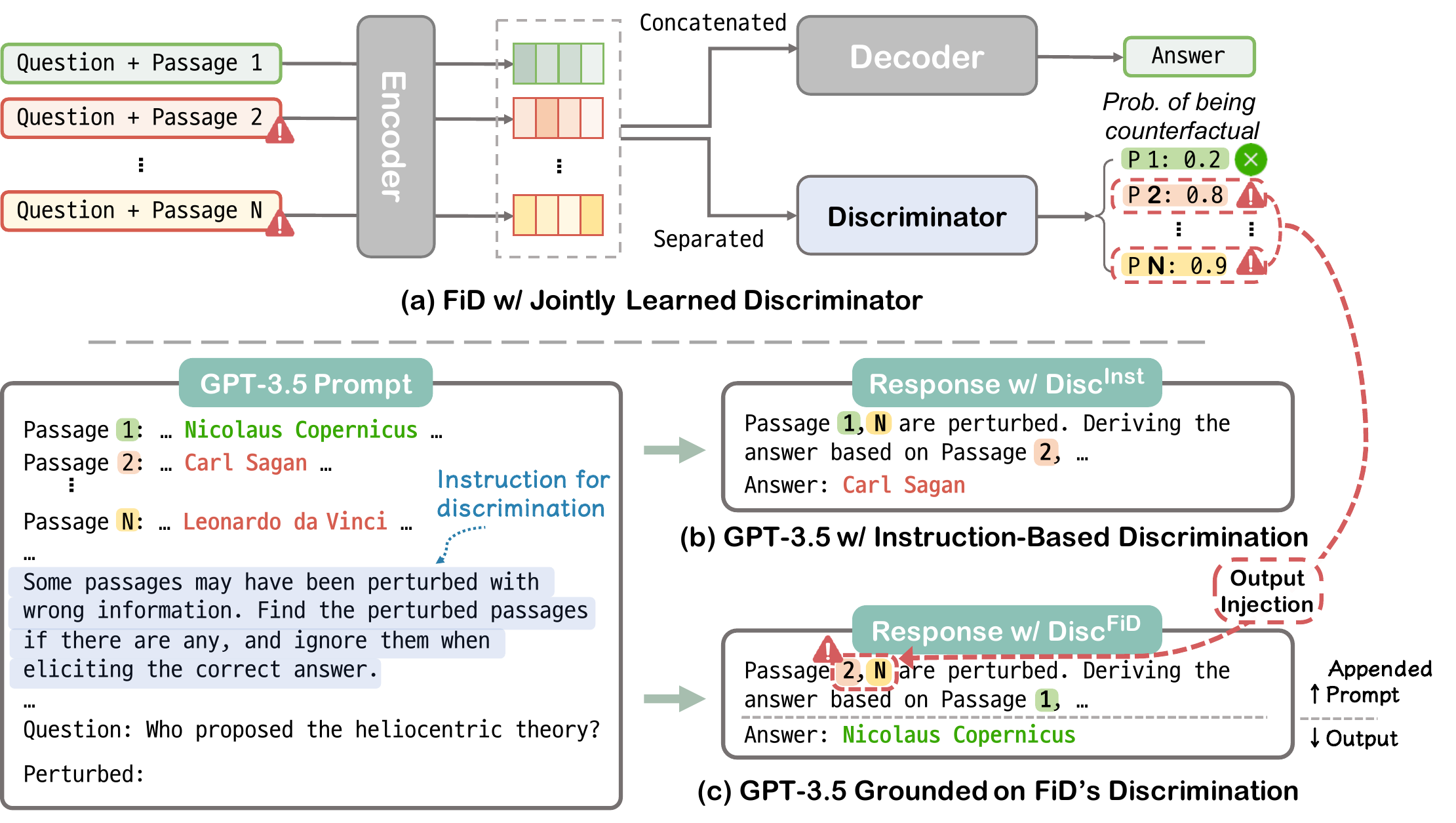}
    \caption{Illustration of our approaches to enhancing robustness to counterfactual noise. \texttt{(a)} Along with the decoder, the discriminator is jointly trained with the downstream task (QA), making the encoder produce corrupt-aware embeddings. \texttt{(b)} GPT-3.5 is prompted to find the perturbed documents before generating an answer. A zero-shot example is shown for brevity. \texttt{(c)} Fine-tuned discriminator output is injected into the prompt for GPT-3.5.}
    \label{fig:main_figure}
\end{figure*}

\section{Related Work}
\paragraph{Retrieval-Augmented Language Models}
Retrieval-augmentation aims to capture world knowledge in a more efficient and interpretable manner \cite{pmlr-v119-guu20a}, and address the hallucination and knowledge update issues \cite{lewis2020retrieval, izacard2021leveraging}. Some works scaled the size of retrieved documents \cite{lakhotia-etal-2021-fid}, while others adopted retrieval to reduce LM's parameter size \cite{pmlr-v162-borgeaud22a}. While promising, most works disregarded the possible prevalence of counterfactual documents.
A recent work \cite{luo2023sail} studies instruction-tuned search-augmentation to filter out distracting documents, motivated by the fact that not all retrieved documents are informative. Our work shares a similar motivation but challenges the binary notion of relevance, as even relevant ones can contain incorrect information, causing conflict.

\paragraph{Knowledge Conflicts and Answer Calibration}
\citet{chen2022rich} and \citet{neeman-etal-2023-disentqa} investigated model behaviors in knowledge conflict settings. They either used calibration \cite{kamath2020selective, zhang-etal-2021-knowing} to abstain from answering, or generated multiple answers upon conflict. Our work, on the contrary, deals with improving the model's ability to distinguish gold from counterfactual information when confronted with knowledge conflicts, providing a correct answer rather than remaining silent. 
\citet{Kazemi2023BoardgameQAAD} argued that available information is frequently inconsistent or contradictory particularly when reasoning in the real-world. They imposed explicit preferences over information sources to resolve conflicts, whereas our approach aims to modulate models' implicit parametric knowledge through discriminator fine-tuning.
A concurrent work, \citet{pan-etal-2023-risk}, studies LLM-generated misinformation. While they use GPT-3.5 to generate documents for explicitly distinct settings, we aim for more natural, challenging, and controllable settings using GPT-4, e.g., introducing the controllability of the noise level (\S \ref{sec:gen_perturb}).
Our method shows stark contrast to their separate fine-tuning and prompting approaches by explicitly combining the intermediate reasoning steps of prompting with the fine-tuned discriminator to detect misinformation (Figure \ref{fig:main_figure} \texttt{(c)}).

\paragraph{Machine-Generated Documents and Misleading Information}
In recent years, machine-generated documents resembling human-written content have raised concerns about misinformation and differentiating their origins from human-written documents \cite{NEURIPS2022_b1efde53}.
For instance, recent work has shown that humans struggle to identify machine-generated writing \cite{clark-etal-2021-thats, kim-etal-2021-distinguish}. The emergence of GPT-4 has further intensified worries about the potential misuse of such models to create deceptive content \cite{openai2023gpt4}. 
Research has revealed that conventional models rarely recognize misinformation but rather contribute to its amplification by generating fabricated details \cite{10.1145/3544548.3581318}. Furthermore, it has been shown that LLM-based applications can be indirectly controlled by adversaries by manipulating retrieval data \cite{greshake2023youve}. These studies motivate the need for robust approaches to address the challenges posed by machine-generated documents. Our work contributes to mitigating the influence of such documents, particularly in the context of retrieval-augmented language models for QA.

\section{Method: \textsc{Discern and Answer}}
We hypothesize that injecting inductive bias \citep{hong-etal-2022-graph, kim-etal-2023-fineprompt} about whether a document may be perturbed or not into a retrieval-augmented LM improves model robustness to conflicting information in QA.
We equip a QA model with a discriminator learned jointly with a QA task, to interpolate the discriminative features with the encoder embeddings so the decoder can capture such a bias when deriving an answer (Figure \ref{fig:main_figure} \texttt{(a)}). Besides fine-tuning, we explore the potential to elicit GPT-3.5's discriminability through in-context instruction, by letting the model explicitly discern before answering (Figure \ref{fig:main_figure} \texttt{(b)}) or injecting fine-tuned model's output into a prompt (Figure \ref{fig:main_figure} \texttt{(c)}).


\subsection{Incorporating Learnable Discriminator into Retrieval-Augmented Model}
\label{sec:disc_train}
Our model builds upon FiD \cite{izacard2021leveraging}, a retrieval-augmented encoder-decoder LM that leverages DPR \cite{karpukhin2020dense} to retrieve a set of $M$ documents from a text corpus $\{d_{1},d_{2}, ...,d_{N}\} \in D$, where $d_{i}$ is retrieved by a similarity search with a question embedding along a document index of size $N$ encoded by a pre-trained BERT \cite{devlin2019bert}. Each document $d_{m}$ is prepended with a question $q$ to be processed independently by a T5 \cite{2020t5} encoder, and is fed to the discriminator (jointly fine-tuned with the encoder).
The discriminator is a one-layer feed-forward network that receives as input each document embedding separately and determines whether the document is perturbed or not; since the information needed to classify a document is encoded by the preceding encoder, a single layer suffices.
The intuition underlying our discriminator fine-tuning is to enhance the encoder's ability to navigate its parametric knowledge space. The resulting encoder representations, therefore, are infused with perturbation-indicative latent information that reduces the influence of perturbed documents on the decoder when it attends over them to generate the final answer.

The encoder representations are concatenated ($\bigparallel$) along the sequence dimension as follows: $H=\bigparallel^{M}_{m=1}Encoder(q, d_{m})$, $H \in \mathbb{R}^{M \times T \times E}$, where $T$ is the maximum sequence length per document and $E$ is the embedding size.

The training objective adopts three complementary loss terms: a generative QA loss $L_{qa}$, a binary cross entropy for discrimination $L_{bce}$, and a contrastive loss $L_{contra}$, formulated as follows:

\small
\setlength{\abovedisplayskip}{3pt}
\begin{align}
    &L_{qa} = {\scriptstyle-}log\ p_{dec}(y|H)\\
    &L_{bce} = \frac{1}{M}\sum^{M}_{m=1}{BCE}(p_{disc}(t_{m}|\boldsymbol{h}^{d_m}), t_{m})\\
    &L_{contra} = {\scriptstyle-}log\frac{\sum_{d^{-} \in \mathcal{D}^{-}_{i}} exp(p_{disc}(t_{m}|\boldsymbol{h}^{d^{-}}))}{\sum_{d^{\pm} \in \mathcal{D}^{+}_{i} \cup \mathcal{D}^{-}_{i}} exp(p_{disc}(t_{m}|\boldsymbol{h}^{d^{\pm}}))}
\label{eq:loss_term}
\end{align}
\normalsize
where $p_{dec}$ and $p_{disc}$ denote the decoder and discriminator probability distribution, respectively. $y$ is the ground-truth answer sequence, $\boldsymbol{h}^{d_m} \in H$ is an encoder representation for the $m$-th document, $t_{m} \in \{0,1\}$ is the perturbation label, $\mathcal{D}^{+}_{i}$ and $\mathcal{D}^{-}_{i}$ are sets of original and perturbed documents, retrieved given the $i$-th question, respectively. 
In essence, these three loss components combined ensure a holistic training signal. 
$L_{qa}$ keeps the primary goal of question answering on track, and $L_{bce}$ retains the encoder's binary classification ability. Inspired by \citet{min-etal-2023-nonparametric}, the adopted $L_{contra}$ considers multiple perturbed and original documents, ensuring that the model does not get overwhelmed by the majority class (i.e., original documents) and continues to learn the adequate nuances of perturbed documents via contrastive objective.
The final loss is $L = L_{qa} + L_{bce} + L_{contra}$. The effects of each term are discussed in \S \ref{subsec:ablation_loss}.

\subsection{Instruction-Based Scheme for Enhancing Robustness to Counterfactual Noise}
\label{prompting}
Our work, in addition to fine-tuning, investigates the effectiveness of instructing GPT-3.5 \cite{NEURIPS2022_b1efde53} to figure out the perturbed documents before answering. Our input prompt consists of (i) a set of retrieved documents partly perturbed by our perturbation scheme in \S \ref{sec:gen_perturb} and \S \ref{sec:gen_perturb_gpt}, followed by (ii) a task-specific instruction (Figure \ref{fig:main_figure} \texttt{(b)}) that prompts the model to explicitly identify and ignore the perturbed documents and generate a correct answer, and (iii) the question that follows afterwards (details are in Figure \ref{fig:prompt_designs} in Appendix \ref{sec:prompt_design}). 

As an extension, we also incorporate the discriminator (\S \ref{sec:disc_train}) to the prompt-based approach. Instead of making GPT-3.5 find the perturbed documents, we insert FiD's discriminator output into the prompt.
This way, we combine the GPT-3.5's rich parametric knowledge and the FiD's task-specific discriminator of high precision (Figure \ref{fig:main_figure} \texttt{(c)}), exhibiting complementarity as discussed in \S \ref{exp_robust}.

\begin{table*}[ht!]
\small
\centering
\begin{adjustbox}{width=1\textwidth}
\begin{tabular}{llccccc}
\toprule
\multirow{2.5}{*}{\shortstack[l]{\textbf{Base}\\\textbf{Model}}} & \multirow{2.5}{*}{\textbf{Method}} & \multicolumn{5}{c}{\textbf{Perturbation \% (Dev / Test)}} \\
\cmidrule(lr){3-7}
 & & \textbf{0\%} & \textbf{15\%} & \textbf{25\%} & \textbf{35\%} & \textbf{Avg.} \\ 
\midrule

\multirow{4}{*}{\shortstack{FiD}} 
 & Parametric (w/o Retrieval) & \multicolumn{4}{c}{12.1 / 14.7} & 12.1 / 14.7 \\
 & Semi-Parametric & \textbf{62.5} / \textbf{63.3} & 44.5 / 47.7 & 41.8 / 40.0 & 28.1 / 30.6 & 44.2 / 45.4 \\
 & Semi-Parametric w/ \texttt{\textbf{$\text{Disc}^{\text{FiD}}$}} & \textbf{62.5} / 63.2 & \textbf{51.6} / \textbf{51.8} & \textbf{43.0} / \textbf{45.6} & \textbf{38.3} / \textbf{36.4} & \textbf{48.9} / \textbf{49.3} \\
 & $\ \ \Delta$ Absolute Gain & \textcolor{lightblue}{+0.0} / \textcolor{lightred}{-0.1} & \textcolor{lightblue}{+7.1} / \textcolor{lightblue}{+4.1} & \textcolor{lightblue}{+1.2} / \textcolor{lightblue}{+5.6} & \textcolor{lightblue}{+10.2} / \textcolor{lightblue}{+5.8} & \textcolor{lightblue}{+4.7} / \textcolor{lightblue}{+3.9} \\
 \midrule
\multirow{5}{*}{\shortstack{GPT-3.5}} 
 & Parametric (w/o Retrieval) & \multicolumn{4}{c}{32.0 / 36.8} & 32.0 / 36.8 \\
 & Semi-Parametric & 50.4 / 53.2 & 40.2 / 45.0 & 31.3 / 37.8 & 22.7 / 24.2 & 36.2 / 40.1 \\
 & Semi-Parametric w/ \texttt{\textbf{$\text{Disc}^{\text{Inst}}$}} & 48.8 / 54.2 & 37.9 / 45.6 & 28.9 / 38.4 & 21.5 / 26.8 & 34.3 / 41.3 \\
 & Semi-parametric w/ \texttt{\textbf{$\text{Disc}^{\text{FiD}}$}} & \textbf{51.2} / \textbf{56.3} & \textbf{42.2} / \textbf{49.2} & \textbf{34.0} / \textbf{41.6} & \textbf{27.3} / \textbf{28.6} & \textbf{38.7} / \textbf{43.9} \\
 & $\ \ \Delta$ Absolute Gain & \textcolor{lightblue}{+0.8} / \textcolor{lightblue}{+3.1} & \textcolor{lightblue}{+2.0} / \textcolor{lightblue}{+4.2} & \textcolor{lightblue}{+2.7} / \textcolor{lightblue}{+3.8} & \textcolor{lightblue}{+4.6} / \textcolor{lightblue}{+4.4} & \textcolor{lightblue}{+2.5} / \textcolor{lightblue}{+3.8} \\
\bottomrule
\end{tabular}
\end{adjustbox}
\caption{Performance in Exact Match (EM) on our \textbf{dev} and \textbf{test} sets (full), according to the perturbation \% of 
retrieved documents. 
GPT-3.5 is ensembled (Appendix \ref{sec:ensemble_results}) over $k=5$ instances (\S \ref{sec:eval_longpre}). $\Delta$ is against Semi-Parametric.}
\label{table:main_results}
\end{table*}

\section{Evaluation under Entity Replacement Framework \cite{longpre-etal-2021-entity}}
\label{sec:eval_longpre}

We measure the performance of FiD and GPT-3.5 (\texttt{text-davinci-003}) in the following settings. The \textbf{Parametric (w/o Retrieval)} setting relies on only rich parametric knowledge \citep{kim-etal-2022-exploiting} to answer a question. The \textbf{Semi-Parametric} setting uses retrieved documents and parametric knowledge; we measure how the infused conflicting information affects the models' performance. Our methods with discrimination (\texttt{\textbf{Disc}}) capabilities are denoted as \textbf{Semi-Parametric + \texttt{Disc}}: the fine-tuned discriminator is superscripted as \texttt{\textbf{$\text{Disc}^{\text{FiD}}$}} and the purely prompt-based discrimination as \texttt{\textbf{$\text{Disc}^{\text{Inst}}$}}.

To fit the maximum length of GPT-3.5, we use the top 5 documents for the dev and test sets for both GPT-3.5 and FiD for a fair comparison. 
Due to the API budget constraint, we sample 256 dev set as in \citet{Le2022FewShotAR}, while using the full test set.
The generated outputs from GPT-3.5 are ensembled over the $k$ instances (Appendix \ref{sec:ensemble_results}) to mitigate the in-context sample sensitivity observed in \citet{pmlr-v139-zhao21c}. Details are in Appendix \ref{sec:fid_train_eval}.

\subsection{Generating Adversarial Documents}
\label{sec:gen_perturb}
Our study explores \textit{the robustness of models under contradictory information, and the influence of varying degrees of noise}.
To facilitate a controllable study, inspired by \citet{kim-etal-2021-seen}, we generate perturbed documents by adopting an entity-centric perturbation strategy \cite{longpre-etal-2021-entity}. This involves taking a document and substituting a gold answer with a randomly sampled named entity of the same type, e.g., Michael Jordan (PER) is replaced with Kobe Bryant (PER). 
We measure the LMs' performance by controlling the proportion of perturbed documents (\textbf{0\%, 15\%, 25\%, 35\%}). 
Details about generation are in Appendix \ref{sec:gen_perturb_detail}.

\subsection{Brittleness of Retrieval-Augmented Models to Conflicting Information}
\label{brittleness}
We analyze how brittle the retrieval-augmented LMs are in the presence of conflict-provoking (i.e., perturbed in the experimental setting) documents for the NQ-Open task \cite{kwiatkowski-etal-2019-natural}. In Table \ref{table:main_results}, we show that the performances of \textbf{Semi-Parametric} for both FiD and GPT-3.5 degrade significantly as the perturbation percentage increases, even when the gold documents are provided. We also note that in a highly perturbed setting (\textbf{35\%}), GPT-3.5's \textbf{Semi-Parametric} becomes worse than its \textbf{Parametric (w/o Retrieval)} counterpart. Our results demonstrate that these seemingly strong models are easily affected by conflicts.

\setlength{\tabcolsep}{2.5pt}
\begin{table}[t!]
\small
\centering
\begin{tabular}{lccccccc} 
\toprule
\multirow{2}{*}{} & \multicolumn{3}{c}{\textbf{FiD}} & \multicolumn{4}{c}{\textbf{GPT-3.5}}\\ 
\cmidrule(lr){2-4}\cmidrule(lr){5-8}
& \textbf{Prec.} & \textbf{Rec.} & \textbf{F1} & & \textbf{Prec.} & \textbf{Rec.} & \textbf{F1} \\
\midrule
 \textbf{15\%} & 93.49 & 61.87 & 74.46 &  & 20.98 & 51.21 & 29.76 \\
 \textbf{25\%} & 95.77 & 64.82 & 77.31 &  & 32.32 & 50.98 & 39.56 \\
 \textbf{35\%} & 97.14 & 69.46 & 81.00 &  & 43.42 & 50.54 & 46.71 \\
\bottomrule
\end{tabular}
\caption{Discriminator performance on our full NQ-Open test set. Each row corresponds to perturbation \%.}
\label{table:disc_performance_test}
\end{table}

\subsection{Improved Robustness via Discriminators}
\label{exp_robust}

For FiD, we see that \textbf{Semi-Parametric w/ \texttt{$\text{Disc}^{\text{FiD}}$}} exhibits improved robustness when confronted with conflicting information (\textbf{15\% - 35\%}), with the average gain of 3.9 on test set. As the proportion of misleading noise increases, there is a general drop in performance while our approach, especially in a highly conflicting scenario (e.g., \textbf{35\%}), exhibits maximum gains. This highlights the discriminator's efficacy in reducing vulnerability to noise.

For GPT-3.5, we observe that \texttt{\textbf{$\text{Disc}^{\text{Inst}}$}} does not incite clear improvement. In Table \ref{table:disc_performance_test}, we show \texttt{\textbf{$\text{Disc}^{\text{Inst}}$}}'s classification performance, where the GPT-3.5's prompt-based few-shot discriminator approach substantially underperforms its fine-tuned counterpart, \texttt{\textbf{$\text{Disc}^{\text{FiD}}$}}. This motivated us to provide \texttt{\textbf{$\text{Disc}^{\text{FiD}}$}}'s output to GPT-3.5 as mentioned in \S \ref{prompting}.
We find this enhances the LLM's robustness in all degrees of noise, highlighting the synergistic interplay between GPT-3.5’s rich parametric knowledge and FiD’s precise task-specific discrimination.

We notice that in \textbf{35\%}, \textbf{Semi-Parametric w/ \texttt{$\text{Disc}^{\text{FiD}}$}} underperforms \textbf{Parametric (w/o Retrieval}) despite the performance recovery from \textbf{Semi-Parametric} (\S \ref{brittleness}).
This is attributed to the high portion of noise caused by the suboptimal recall (Table \ref{table:disc_performance_test}), which is exacerbated by GPT-3.5's strong prompt-following characteristics \cite{NEURIPS2022_b1efde53} as evidenced by \textbf{Semi-Parametric}'s high susceptibility in \S \ref{brittleness}. This indicates room for further improvement in future work.

\begin{table*}[ht!]
\small
\centering
\begin{adjustbox}{width=0.95\textwidth}
\setlength{\tabcolsep}{6pt}
\begin{tabular}{lcccccccc}
\toprule
 \multirow{2.5}{*}{\textbf{Method}} & \multicolumn{4}{c}{\textbf{QA (EM)}} & \multicolumn{4}{c}{\textbf{Classification (F1)}} \\ 
 \cmidrule(lr){2-5}\cmidrule(lr){6-9}
 & \textbf{15\%} & \textbf{25\%} & \textbf{35\%} & \textbf{Avg.} & \textbf{15\%} & \textbf{25\%} & \textbf{35\%} & \textbf{Avg.} \\
\midrule
Semi-parametric & 44.53 & 41.80 & 28.12 & 38.15 & - & - & - & - \\
\quad+ Disc. ($L_{qa} + L_{bce}$) & 49.22 & 43.75 & 35.94 & 42.97 & 73.48 & 75.77 & \textbf{82.65} & \textbf{77.30} \\
\quad+ Disc. ($L_{qa} + L_{contra}$) & 45.70 & \textbf{44.92} & 37.11 & 42.58 & 59.47 & 71.02 & 74.91 & 68.47 \\
\quad+ Disc. ($L_{qa} + L_{bce} + L_{contra}$) & \textbf{51.56} & 42.97 & \textbf{38.28} & \textbf{44.27} & \textbf{74.05 }& \textbf{77.43} & 80.15 & 77.21 \\
\bottomrule
\end{tabular}
\end{adjustbox}
\caption{Ablation study on the loss terms.}
\label{table:loss_term}
\end{table*}

\subsection{Enhanced In-Context Learning Stability}
\label{subsec:enhanced_icl_stability}
\begin{figure}[h!]
    \centering
    \includegraphics[scale=0.41]{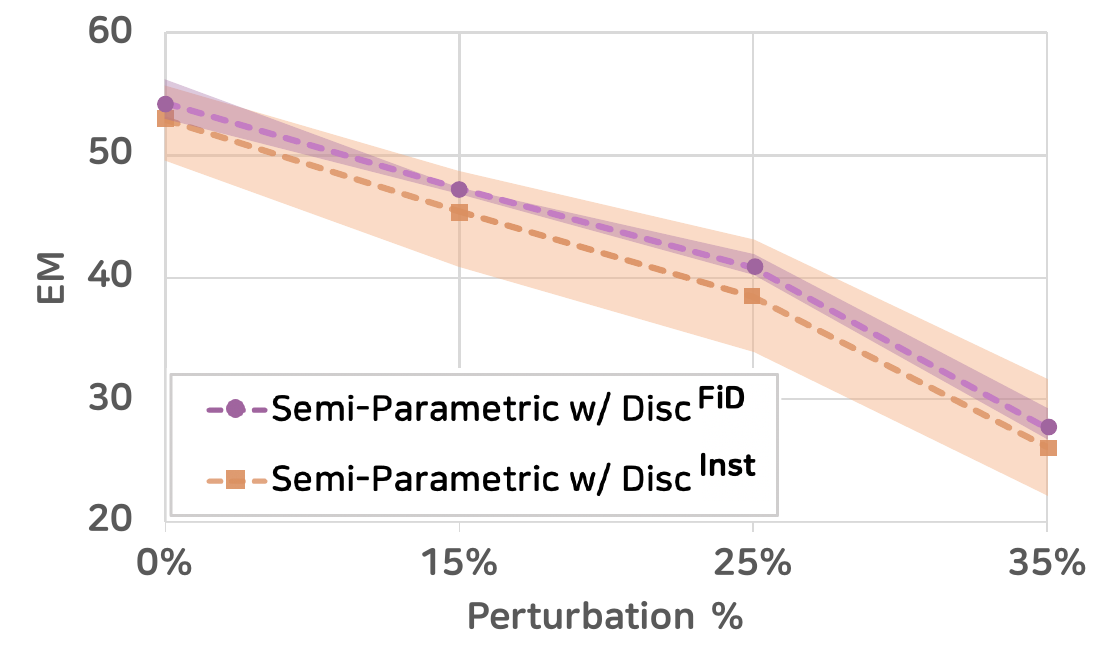}
    \caption{Comparison of GPT-3.5's stability for each discriminator setting.
    The shaded area represents the variance computed between the best and worst EM.} \label{fig:stability_figure}
\end{figure}
Figure \ref{fig:stability_figure} shows the best, average and worst EM scores of GPT-3.5 over 5 different in-context samples. In-context learning is known for its high instability \cite{pmlr-v139-zhao21c, min-etal-2022-noisy}, and we discover that injecting the fine-tuned discriminator into the in-context learning (\textbf{GPT-3.5 (Semi-parametric w/ \texttt{\textbf{$\text{Disc}^{\text{FiD}}$}})}) greatly improves the stability. This new facet along with the result in \S \ref{exp_robust}, which shows complementarity, highlights the potential of leveraging both strengths of fine-tuning and in-context learning paradigms.

\subsection{Ablation Study}
\label{subsec:ablation_loss}
To demonstrate the effect of different loss terms in fine-tuning our discriminator, we provide the results of our ablation study in Table \ref{table:loss_term}. The simple binary classification loss, $L_{bce}$, which is jointly minimized with the QA loss, $L_{qa}$, markedly improves performance in the perturbed scenarios. We also evaluate the contrastive objective, $L_{contra}$ between perturbed and original documents. While the sole addition of $L_{contra}$ underperforms both the QA and perturbation classification, we show that it shares a complementary relationship with $L_{bce}$, greatly improving the overall performance across different perturbation configurations; we therefore select this setting as our proposed model.

\subsection{Task Transferability to TriviaQA}
\label{sec:task_transfer_tqa}

\begin{figure}[th!]
    \centering
    \includegraphics[scale=0.26]{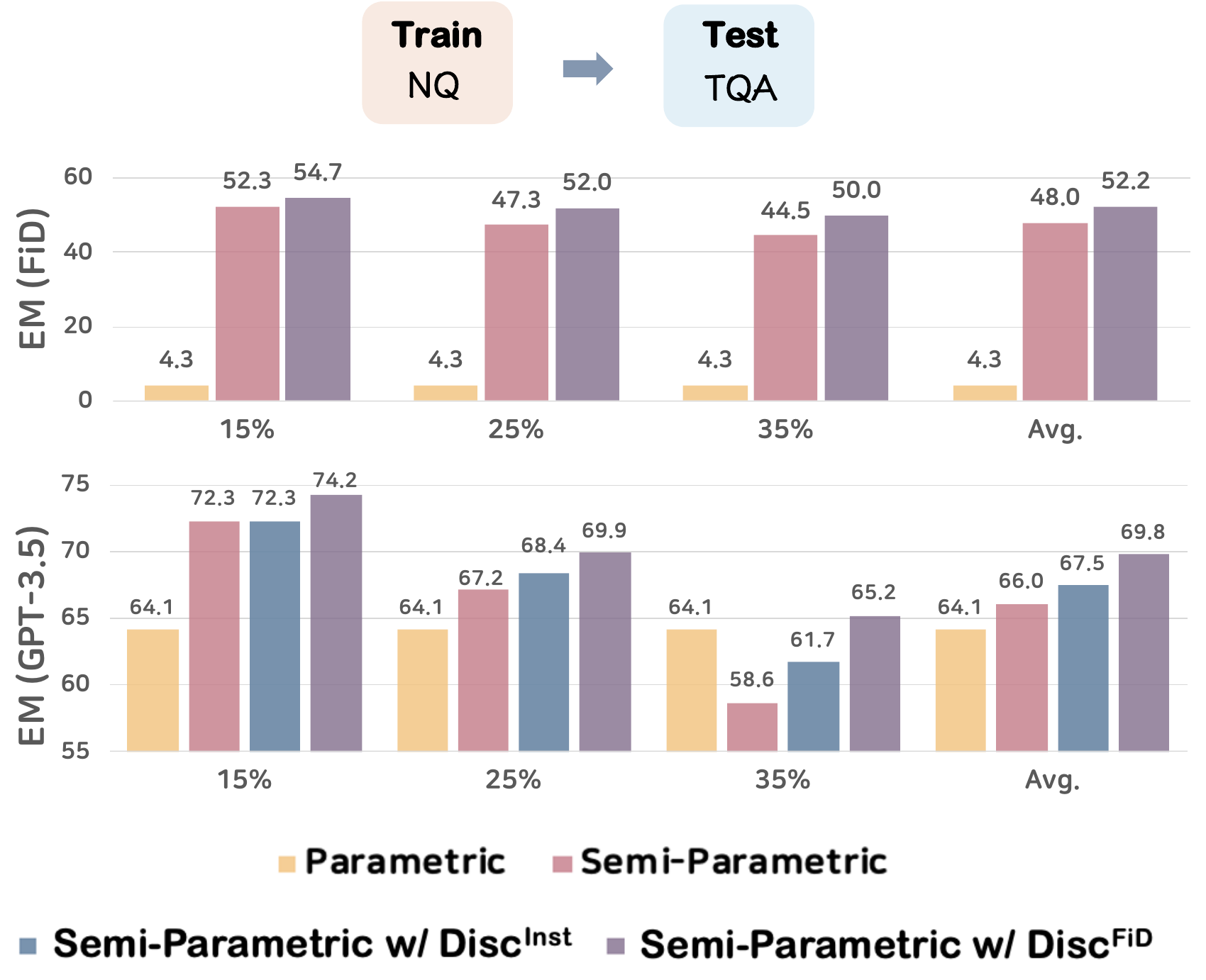}
    \caption{Results on TQA-open \textbf{dev}. FiD (i.e., discriminator) is trained on NQ-open and evaluated on TQA-open to examine the transferability of the robustness acquired through our method.} \label{fig:triviaqa}
\end{figure}

While our models demonstrate promising results on NQ-open, it remains questionable whether the models can generalize to other datasets.
To evaluate the transferability and robustness of our fine-tuned discriminator on other related tasks, we evaluated the performance of our models on the TriviaQA (TQA-open) \cite{joshi-etal-2017-triviaqa} dev set as shown in Figure \ref{fig:triviaqa}; the discriminator was fine-tuned only on the NQ-open dataset. The results show that the discriminator is able to distinguish perturbed documents from original ones given the performance gains on the perturbed TQA-open dataset. This suggests that our fine-tuned discriminator, even when it is not explicitly fine-tuned on an end task dataset, is able to extend its discriminability to other tasks. Furthermore, the retention of robustness in the perturbed TQA-open setting serves as a testament that our discriminator does not rely on shortcuts or memorization to distinguish perturbed documents. Test set results exhibiting similar trends are shown in Figure \ref{fig:triviaqa_test} in the Appendix \ref{subsec:addl_result_entpert}.

\begin{table*}[th!]
{\renewcommand{\arraystretch}{1.2}
\small
\tabcolsep=0.1cm\centering
\begin{adjustbox}{width=0.95\textwidth}
\begin{tabular}[t]{ lccccc } 
\toprule
     & \shortstack[c]{\textbf{Context}\\\textbf{Mismatch}} & \shortstack[c]{\textbf{Question}\\\textbf{Answerability}} & \shortstack[c]{\textbf{Document}\\\textbf{Length}} & \shortstack[c]{\textbf{Counter-}\\\textbf{factuality}} & \shortstack[c]{\textbf{Perturbation}\\ \textbf{Type}} \\
\midrule\midrule
  \textbf{Entity Replacement} & 27.5\% & 100.0\% & 106 & 100.0\% & ER (100.0\%) \\
  \textbf{\textsc{MacNoise}} & 0.0\% & 100.0\% & 123 & 91.8\% & AC (8.9\%) | GR (21.9\%) | LR (45.2\%) | ER (24.0\%) \\
\bottomrule
\end{tabular}
\end{adjustbox}
\caption{Comparison between entity replacement framework vs. our \textsc{MacNoise}. AC: Additional Context. GR: Global Revision. LR: Local Revision. ER: Entity Replacement w/ Context Match.}
\label{table:data_analysis}
}
\end{table*}

\section{Evaluation on New Machine-Generated Noise (\textsc{MacNoise}) Benchmark}
\label{sec:eval_gpt4}

To extend our evaluation scope beyond the entity replacement, we present \textsc{MacNoise}, a \textbf{Mac}hine-Generated \textbf{Noise} dataset for ODQA containing knowledge conflicts among evidence documents. \textsc{MacNoise} aims to provide more realistic knowledge conflict scenarios compared to the previous entity-centric perturbation framework, addressing limitations discussed in the subsequent section.

\subsection{Limitations of Entity Perturbation Framework \cite{longpre-etal-2021-entity}}
\label{subsec:limitations_entity_framework}

While the widely-used entity replacement framework (\S \ref{sec:eval_longpre}) serves as a simple and scalable proxy for understanding the knowledge conflict scenario in ODQA setting, we posit that, intuitively, the perturbed documents may exhibit the following potential issues: 
\begin{itemize}
    \itemsep0em
    \item Context mismatch: the replaced entities may not be aligned with the co-occurring context (e.g., "Victoria's Secret was founded by \textit{Roy \textcolor{green}{Raymond}}, and to his wife Gaye \textcolor{green}{Raymond}" to "Victoria's Secret was founded by \textit{Patrick \textcolor{red}{Denham}}, and to his wife Gaye \textcolor{red}{Raymond}) and this may also entail pronoun mismatch.
    \item Confined noise type: the perturbation scheme focuses only on removing the existing answer entity from the input passage; it does not employ any other alternative noise generation strategy (e.g., answer negation, multiple answers) that helps enhance the verisimilitude of the documents.
    \item Semantic equivalence: with low probability, semantically equivalent entities such as aliases may be put in place of the original answer entity within the context (e.g., "The author \textit{Samuel Clemens} wrote `The Adventures of Tom Sawyer'" to "The author \textit{Mark Twain} wrote `The Adventures of Tom Sawyer'").
\end{itemize}


These implausible cases risk the manifestation of shortcuts within the models trained on the entity-swapped documents. As such, we introduce \textsc{MacNoise} (\S \ref{sec:gen_perturb_gpt}) to mitigate the model’s reliance on these synthetic cues.
Since the three problems with \citet{longpre-etal-2021-entity} may cause robustness issues in our proposed system in a more realistic environment, we fine-tune our \textbf{Semi-Parametric + \texttt{\textbf{$\text{Disc}^{\text{FiD}}$}}} on LLM-generated knowledge conflict documents (\S \ref{sec:gen_perturb_gpt}) from \texttt{GPT-3.5-turbo}. To see if our fine-tuned model can fend off a more challenging machine-generated noise among retrieved documents, we generate our evaluation dataset with GPT-4 \cite{openai2023gpt4}, the most powerful existing LLM in both commercial and open-source domains.
Our dataset generation's significance is highlighted by the prevalence of machine-generated noise \cite{openai2023gpt4, 10.1145/3544548.3581318} due to the growing usage of LLMs in general. The notorious hallucination issue inundates the Web environment with noisy, potentially fallacious texts, creating a hazardous environment for retrieval-augmented LMs to exploit knowledge from - another cause for our additional dataset generation. We provide the actual prompt used to generate our dataset using the LLMs in Table \ref{table:macnoise_prompt} in Appendix \ref{subsec:generate_macnoise}.

\begin{table*}[ht!]
\centering
\begin{adjustbox}{width=0.9\textwidth}
\setlength{\tabcolsep}{6pt}
\begin{tabular}{llccccccccccc}
\toprule
\multirow{2.5}{*}{\shortstack[l]{\textbf{Base}\\\textbf{Model}}} & \multirow{2.5}{*}{\textbf{Method}} & \multicolumn{5}{c}{\textbf{Perturbation \% (NQ-open)}} & & \multicolumn{5}{c}{\textbf{Perturbation \% (TQA-open)}} \\
\cmidrule(lr){3-7}\cmidrule(lr){9-13}
 & & \textbf{0\%} & \textbf{15\%} & \textbf{25\%} & \textbf{35\%} & \textbf{Avg.}
 & & \textbf{0\%} & \textbf{15\%} & \textbf{25\%} & \textbf{35\%} & \textbf{Avg.}
 \\ 
\midrule

\multirow{4}{*}{\shortstack{FiD}} 
 & Parametric (w/o Retrieval) & \multicolumn{4}{c}{12.1} & 12.1 & & \multicolumn{4}{c}{4.3} & 4.3 \\
 & Semi-Parametric & \textbf{62.5} & 50.8 & 39.1 & 28.5 & 45.2 & & \textbf{61.7} & 54.3 & 48.8 & 35.9 & 50.2 \\
 & Semi-Parametric w/ \texttt{\textbf{$\text{Disc}^{\text{FiD}}$}} & \textbf{62.5} & \textbf{52.0} & \textbf{41.4} & \textbf{30.1} & \textbf{46.5} & & 60.9 & \textbf{60.6} & \textbf{53.5} & \textbf{48.1} & \textbf{55.8}  \\
 & $\ \ \Delta$ Absolute Gain & \textcolor{lightblue}{+0.0} & \textcolor{lightblue}{+1.2} & \textcolor{lightblue}{+2.3} & \textcolor{lightblue}{+1.6} & \textcolor{lightblue}{+1.3} & & \textcolor{lightred}{-0.8} & \textcolor{lightblue}{+6.3} & \textcolor{lightblue}{+4.7} & \textcolor{lightblue}{+12.2} & \textcolor{lightblue}{+5.6} \\
 \midrule
\multirow{5}{*}{\shortstack{GPT-3.5}} 
 & Parametric (w/o Retrieval) & \multicolumn{4}{c}{32.0} & 32.0 & & \multicolumn{4}{c}{64.1} & 64.1 \\
 & Semi-Parametric & 50.4 & 28.5 & 23.8 & 16.0 & 29.7 & & 71.9 & 60.9 & 53.5 & 43.0 & 57.3 \\
 & Semi-Parametric w/ \texttt{\textbf{$\text{Disc}^{\text{Inst}}$}} & 48.8 & 36.3 & 28.5 & 19.5 & 33.3 & & 73.8 & 64.1 & 56.6 & 44.9 & 59.9 \\
 & Semi-parametric w/ \texttt{\textbf{$\text{Disc}^{\text{FiD}}$}} & \textbf{51.2} & \textbf{37.1} & \textbf{30.1} & \textbf{21.5} & \textbf{35.0} & & \textbf{76.2} & \textbf{68.0} & \textbf{61.7} & \textbf{53.1} & \textbf{64.7} \\
 & $\ \ \Delta$ Absolute Gain & \textcolor{lightblue}{+0.8} & \textcolor{lightblue}{+8.6} & \textcolor{lightblue}{+6.3} & \textcolor{lightblue}{+5.5} & \textcolor{lightblue}{+5.3} & & \textcolor{lightblue}{+4.3} & \textcolor{lightblue}{+7.1} & \textcolor{lightblue}{+8.2} & \textcolor{lightblue}{+10.1} & \textcolor{lightblue}{+7.4}\\
\bottomrule
\end{tabular}
\end{adjustbox}
\caption{Performance in Exact Match (EM) on our dev of NQ-open and TQA-open w/ machine-generated conflict (\textsc{MacNoise}), according to the perturbation \% of 
retrieved documents. 
GPT-3.5 is ensembled (Appendix \ref{sec:ensemble_results}) over $k=5$ instances (\S \ref{sec:eval_longpre}). $\Delta$ is against Semi-Parametric.}
\label{table:main_results_llm}
\end{table*}

\setlength{\tabcolsep}{2.5pt}
\begin{table}[t]
\small
\centering
\begin{tabular}{lccccccc} 
\toprule
\multirow{2}{*}{} & \multicolumn{3}{c}{\textbf{FiD}} & \multicolumn{4}{c}{\textbf{GPT-3.5}}\\ 
\cmidrule(lr){2-4}\cmidrule(lr){5-8}
& \textbf{Prec.} & \textbf{Rec.} & \textbf{F1} & & \textbf{Prec.} & \textbf{Rec.} & \textbf{F1} \\
\midrule
 \textbf{15\%} & 97.58 & 63.35 & 76.83 &  & 17.72 & 50.89 & 25.74 \\
 \textbf{25\%} & 96.57 & 63.14 & 76.36 &  & 26.13 & 49.94 & 34.31 \\
 \textbf{35\%} & 96.32 & 69.32 & 80.62 &  & 37.94 & 50.91 & 43.48 \\
\bottomrule
\end{tabular}
\caption{Classification performance of our discriminator on the NQ-open with \textsc{MacNoise}.}
\label{table:disc_performance_test_machine}
\end{table}

\subsection{Generating Counterfactual Documents using Large Language Models}
\label{sec:gen_perturb_gpt}
Using GPT-4 and \texttt{GPT-3.5-turbo}, we generate our evaluation and training datasets, respectively (dataset generation details in Appendix \ref{subsec:generate_macnoise}). 
To address the limitations of the entity-perturbed documents, we leverage the fact that LLM-generated texts are indistinguishable \cite{clark-etal-2021-thats} from human-generated texts, and LLMs closely adhere to the given instructions, in which our dataset generation constraints are given. We elaborate on the instruction formulation in this section.

\paragraph{Perturbation Instruction}
Our perturbation instruction constrains the LLMs with the following rules when generating noise-injected documents: 
(i) \textit{Question answerability} - perturbed documents should be answerable with the paired question; any information requested by the question can be changed but the documents should retain their relevance to the question. 
(ii) \textit{Length similarity} - perturbed documents should be similar in length to the original document. We impose this constraint to address GPT-model's notorious tendency to generate verbose texts \cite{liu2023sociallyaligned}.
(iii) \textit{Answer Perturbation} - the model should either remove the original answer span or revise the document so the context no longer supports the answer.

We also provide the LLMs with a set of revision strategies to create the perturbed documents. The revision strategies are similar to rule (iii), prompting the model to rewrite the document so the document no longer supports the answer, to replace the entities in the passage, or to negate the sentences the answer span appears in so that the original answer span no longer supports the answer. The actual instruction used is described in Appendix \ref{sec:gen_perturb_detail}.

\paragraph{Comparison to Entity-Perturbed Documents} 

Here, we provide both the quantitative and qualitative comparison of LLM-generated against entity-perturbed documents; the LLM used here is \mbox{GPT-4} \cite{openai2023gpt4}. In Table \ref{table:data_analysis}, we demonstrate that the LLM-generated documents adequately address the problems of the entity-based perturbation scheme used in \S \ref{sec:eval_longpre} while retaining their similarity to the original documents in terms of context length and answer validity rate. We sample 64 instances from the GPT-4-generated dev set of the NQ-open dataset; this consists of a total of 320 documents wherein 146 documents (44.24\%) are perturbable. 

Through manual analysis on the sampled documents, we identified four perturbation types that distinguish the LLM-generated documents from the entity-perturbed ones: (i) Additional Context - most of the original context is retained while the answers are replaced along with a few additional sentences that justify the replaced entity, which explains the slight increase in \textbf{Context Length} in Table \ref{table:data_analysis}; (ii) Global Revision - the entire context of a document is largely rewritten by the LLM ; (iii) Local Revision - the original context is largely retained while the answers are replaced with minor edits in the given context; (iv) Entity Replacement w/ Context Match - this is analogous to \citet{longpre-etal-2021-entity} while avoiding context mismatch. 
Refer to Appendix \ref{sec:gen_perturb_detail} for dataset statistics and Table \ref{table:case_study} and \ref{table:case_study_ac} in Appendix \ref{subsec:case_study} for case study.

\subsection{Brittleness and Enhanced Robustness to LLM-Generated Conflicts}
We now benchmark models on our LLM-generated conflicts (\textsc{MacNoise}). Note that the perturbed documents used for evaluation are generated using a more powerful GPT-4 \cite{openai2023gpt4}, posing a more challenging scenario for our discriminator, which is fine-tuned on a dataset perturbed by \texttt{GPT-3.5-turbo}.
In Table \ref{table:main_results_llm}, we note an even greater drop (e.g., 50.4 in 0\% $\rightarrow$ 16.0 in 35\% on NQ) for Semi-Parametric GPT-3.5 (\texttt{text-davinci-003}) when confronted with our adversarially generated documents, compared to the entity-perturbed ones (50.4 in 0\% $\rightarrow$ 22.7 in 35\%) in Table \ref{table:main_results}. This observation not only exposes the vulnerability of existing models, but also underscores the fact that our \textsc{MacNoise} benchmark is challenging.
Meanwhile, our fine-tuned discriminator enhances the robustness of both models to LLM-generated perturbed documents. 
In particular, we demonstrate GPT-3.5's over-reliance on the retrieved documents, containing counterfactual texts, can be alleviated to better distinguish perturbed documents, leading to more accurate answers.

\subsection{Complementarity of Entity Replacement and LLM-Generated Perturbations}
\label{subsec:complementarity}
\setlength{\fboxsep}{0pt}
\begin{figure}[th!]
    \centering
    \includegraphics[scale=0.27]{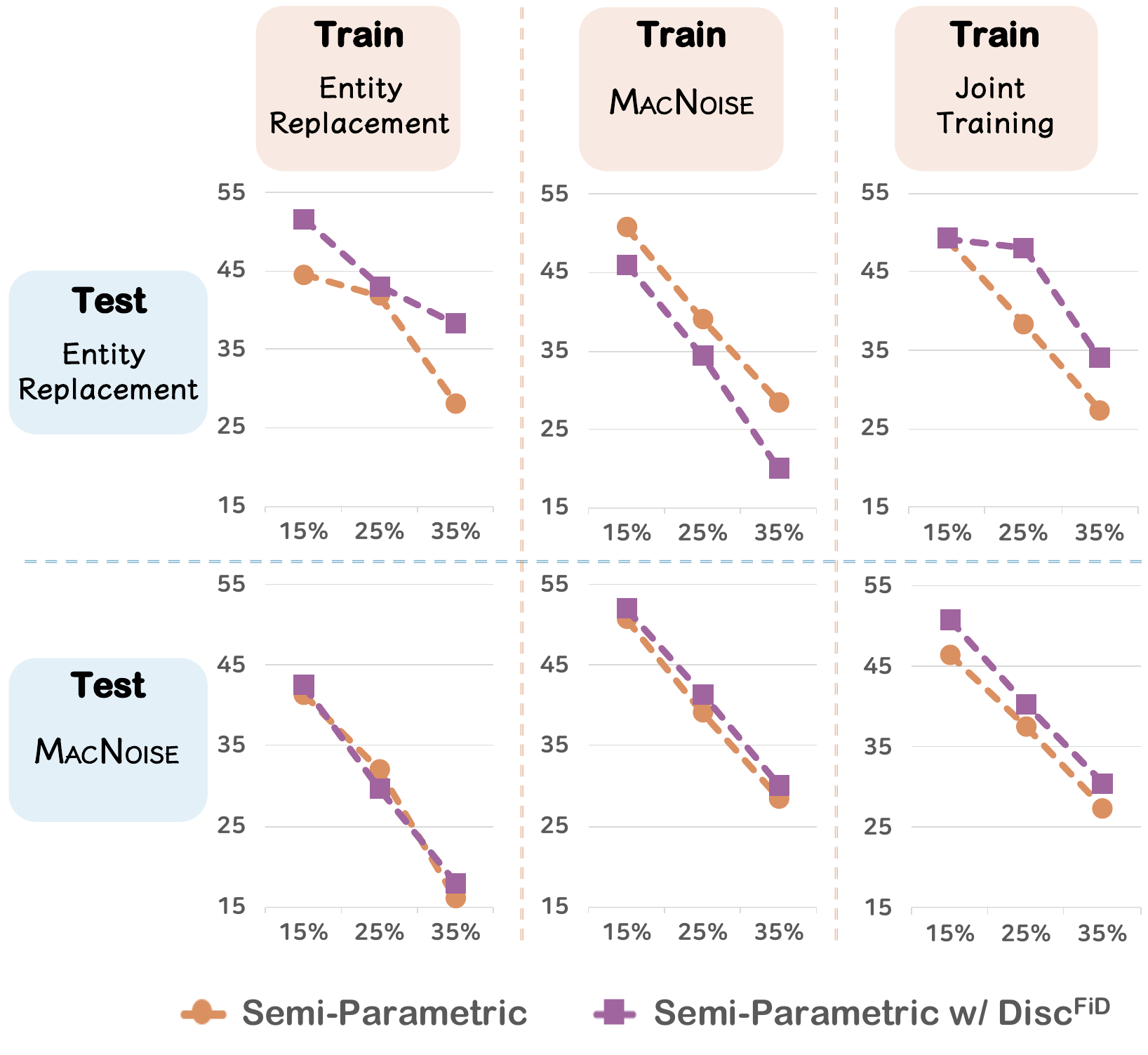}
    \caption{EM scores of the Semi-Parametric and our Semi-Parametric w/ \texttt{\textbf{$\text{Disc}^{\text{FiD}}$}} on the NQ-open \textbf{dev} w/ different perturbations: \colorbox{test}{\texttt{Entity Replacement}} or \colorbox{test}{\textsc{MacNoise}}. The discriminator is fine-tuned independently (either w/ \colorbox{train}{\texttt{Entity Replacement}} or \colorbox{train}{\textsc{MacNoise}}) or jointly (\colorbox{train}{\texttt{Joint Training}}) on the NQ-open \textbf{train}.} \label{fig:joint}
\end{figure}
As an additional experiment, we also evaluate whether the different characteristics of entity-perturbed and LLM-perturbed documents learned during the fine-tuning can be transferred to one another. After jointly training our discriminator with the entity-perturbed (\S \ref{sec:eval_longpre}) and \textsc{MacNoise} (\S \ref{sec:eval_gpt4}) datasets, we can see that the discriminator is able to address the counterfactual noise in both the entity- and LLM-perturbed settings simultaneously (Figure \ref{fig:joint}). This suggests that dealing with different kinds of perturbations simultaneously requires jointly training over the different perturbed document sets, which highlights the importance of curating both the entity- and LLM-perturbed datasets for fine-tuning our discriminator. 

\section{Conclusion}
This work investigates the robustness of retrieval-augmented LMs when the retrieved documents include conflicting information. We show that (i) both the fine-tuned LMs and in-context learned LLMs are brittle to the presence of misleading information, and (ii) our perturbation discriminating approach significantly enhances the LMs' ability to handle conflicts. Furthermore, we find that (iii) combining the fine-tuned discriminator's output with in-context learning improves the LLMs' stability and robustness, creating a new avenue for future work to utilize the advantages of both learning paradigms. We also release \textsc{MacNoise}, an LLM-generated knowledge conflict dataset for ODQA, to facilitate further research.

\section*{Limitations}

In the following, we discuss the limitations of our work to encourage future efforts.

\paragraph{Incurred Costs and Data Sampling}
The use of GPT-3.5 (\texttt{text-davinci-003}) for in-context learning (\S \ref{prompting}) incurs substantial cost because of its price (\$0.02 per 1,000 tokens). Also, the process of creating our \textsc{MacNoise} (\S \ref{sec:gen_perturb_gpt}) also incurs additional costs because \texttt{GPT-3.5-turbo} (\$0.002 per 1,000 output tokens) and GPT-4 (\$0.06 per 1,000 output tokens) are used to generate our training and evaluation datasets, respectively. To accommodate our budget constraints, we sample 256 instances \cite{Le2022FewShotAR} from both the NQ-open and TQA-open dev sets. Nonetheless, our results in Tables \ref{table:main_results} and \ref{table:main_results_llm} clearly demonstrate the efficacy of our proposed fine-tuned discriminator and prompting approaches.

\paragraph{Maximum Input Length of GPT-3.5}
Moreover, to fit the maximum input length of GPT-3.5, we use the top 5 documents for the dev and test sets for our baselines GPT-3.5 and FiD to facilitate a fair comparison. 
We also note that the availability or capability of certain models that need to be accessed through APIs, such as GPT-4 may be subject to change over time.

\paragraph{LLM-Generated Nature of \textsc{MacNoise} Benchmark}
\textsc{MacNoise} is meant to address the synthetic nature of the previous framework \citep{longpre-etal-2021-entity}, in which models may learn to exploit shortcuts to identify misinformation. 
While emulating more realistic counterfactual documents via \textsc{MacNoise}, we acknowledge the inherent nature of the LLM-generated data, which can still be deemed synthetic and artifactual \citep{kang-etal-2020-regularization, hong-etal-2020-handling, das2024under}.

\paragraph{Additional Robustness to Counterfactuals}
Ideally, our fine-tuned discriminator framework should completely suppress the influence of counterfactual information among retrieved documents for FiD and GPT-3.5. While our method substantially improves the performance of these models with our fine-tuned discriminator when the counterfactual information is present in the retrieved documents, the models are nonetheless influenced by the perturbed documents. We encourage future works to further mitigate the influence of counterfactual information for more robust retrieval-augmented generation in language models.

\section*{Ethics Statement}
Our work deals with improving the robustness of retrieval-augmented LMs when conflicting information is present among the retrieved documents. To emulate the scenario, our work purposefully, without any ill-intention, perturbed the retrieved documents with the entity-perturbation framework adopted from a previous work \cite{longpre-etal-2021-entity} and our LLM-generated \textsc{MacNoise} dataset. Importantly, our goal is to address the issue of misleading information in the ODQA setting. 
During the validation process for \textsc{MacNoise}, as presented in Table \ref{table:data_analysis} and elaborated in \S \ref{sec:gen_perturb_gpt}, we diligently screened our dataset to ensure the absence of offensive content or personal information. Moreover, given our utilization of GPT-4 for generation, we acknowledge the privacy considerations highlighted in GPT-4 technical report \cite{openai2023gpt4}; The model has been trained on a diverse set of licensed, created, and publicly available data, some of which might encompass publicly available personal information. Nevertheless, stringent steps have been taken to mitigate the potential risks associated with privacy issues.

\section*{Acknowledgements}
This work was supported by Institute for Information \& communications Technology Planning \& Evaluation(IITP) grant funded by the Korea government(MSIT) (No. 2022-0-00369, (Part 4) Development of AI Technology to support Expert Decision-making that can Explain the Reasons/Grounds for Judgment Results based on Expert Knowledge).

\bibliography{anthology,custom}

\begin{thebibliography}{44}
\expandafter\ifx\csname natexlab\endcsname\relax\def\natexlab#1{#1}\fi

\bibitem[{Borgeaud et~al.(2022)Borgeaud, Mensch, Hoffmann, Cai, Rutherford,
  Millican, Van Den~Driessche, Lespiau, Damoc, Clark, De~Las~Casas, Guy,
  Menick, Ring, Hennigan, Huang, Maggiore, Jones, Cassirer, Brock, Paganini,
  Irving, Vinyals, Osindero, Simonyan, Rae, Elsen, and
  Sifre}]{pmlr-v162-borgeaud22a}
Sebastian Borgeaud, Arthur Mensch, Jordan Hoffmann, Trevor Cai, Eliza
  Rutherford, Katie Millican, George~Bm Van Den~Driessche, Jean-Baptiste
  Lespiau, Bogdan Damoc, Aidan Clark, Diego De~Las~Casas, Aurelia Guy, Jacob
  Menick, Roman Ring, Tom Hennigan, Saffron Huang, Loren Maggiore, Chris Jones,
  Albin Cassirer, Andy Brock, Michela Paganini, Geoffrey Irving, Oriol Vinyals,
  Simon Osindero, Karen Simonyan, Jack Rae, Erich Elsen, and Laurent Sifre.
  2022.
\newblock \href {https://proceedings.mlr.press/v162/borgeaud22a.html}
  {Improving language models by retrieving from trillions of tokens}.
\newblock In \emph{Proceedings of the 39th International Conference on Machine
  Learning}, volume 162 of \emph{Proceedings of Machine Learning Research},
  pages 2206--2240. PMLR.

\bibitem[{Brown et~al.(2020)Brown, Mann, Ryder, Subbiah, Kaplan, Dhariwal,
  Neelakantan, Shyam, Sastry, Askell, Agarwal, Herbert-Voss, Krueger, Henighan,
  Child, Ramesh, Ziegler, Wu, Winter, Hesse, Chen, Sigler, Litwin, Gray, Chess,
  Clark, Berner, McCandlish, Radford, Sutskever, and
  Amodei}]{NEURIPS2020_1457c0d6}
Tom Brown, Benjamin Mann, Nick Ryder, Melanie Subbiah, Jared~D Kaplan, Prafulla
  Dhariwal, Arvind Neelakantan, Pranav Shyam, Girish Sastry, Amanda Askell,
  Sandhini Agarwal, Ariel Herbert-Voss, Gretchen Krueger, Tom Henighan, Rewon
  Child, Aditya Ramesh, Daniel Ziegler, Jeffrey Wu, Clemens Winter, Chris
  Hesse, Mark Chen, Eric Sigler, Mateusz Litwin, Scott Gray, Benjamin Chess,
  Jack Clark, Christopher Berner, Sam McCandlish, Alec Radford, Ilya Sutskever,
  and Dario Amodei. 2020.
\newblock \href
  {https://proceedings.neurips.cc/paper/2020/file/1457c0d6bfcb4967418bfb8ac142f64a-Paper.pdf}
  {Language models are few-shot learners}.
\newblock In \emph{Advances in Neural Information Processing Systems},
  volume~33, pages 1877--1901. Curran Associates, Inc.

\bibitem[{Chen et~al.(2022)Chen, Zhang, and Choi}]{chen2022rich}
Hung-Ting Chen, Michael~JQ Zhang, and Eunsol Choi. 2022.
\newblock Rich knowledge sources bring complex knowledge conflicts:
  Recalibrating models to reflect conflicting evidence.
\newblock \emph{arXiv preprint arXiv:2210.13701}.

\bibitem[{Clark et~al.(2021)Clark, August, Serrano, Haduong, Gururangan, and
  Smith}]{clark-etal-2021-thats}
Elizabeth Clark, Tal August, Sofia Serrano, Nikita Haduong, Suchin Gururangan,
  and Noah~A. Smith. 2021.
\newblock \href {https://doi.org/10.18653/v1/2021.acl-long.565} {All that{'}s
  {`}human{'} is not gold: Evaluating human evaluation of generated text}.
\newblock In \emph{Proceedings of the 59th Annual Meeting of the Association
  for Computational Linguistics and the 11th International Joint Conference on
  Natural Language Processing (Volume 1: Long Papers)}, pages 7282--7296,
  Online. Association for Computational Linguistics.

\bibitem[{Das et~al.(2024)Das, De~Langis, Martin, Kim, Lee, Kim, Hayati, Owan,
  Hu, Parkar et~al.}]{das2024under}
Debarati Das, Karin De~Langis, Anna Martin, Jaehyung Kim, Minhwa Lee, Zae~Myung
  Kim, Shirley Hayati, Risako Owan, Bin Hu, Ritik Parkar, et~al. 2024.
\newblock Under the surface: Tracking the artifactuality of llm-generated data.
\newblock \emph{arXiv preprint arXiv:2401.14698}.

\bibitem[{Devlin et~al.(2019)Devlin, Chang, Lee, and
  Toutanova}]{devlin2019bert}
Jacob Devlin, Ming-Wei Chang, Kenton Lee, and Kristina Toutanova. 2019.
\newblock Bert: Pre-training of deep bidirectional transformers for language
  understanding.
\newblock In \emph{Proceedings of the 2019 Conference of the North American
  Chapter of the Association for Computational Linguistics: Human Language
  Technologies, Volume 1 (Long and Short Papers)}, pages 4171--4186.

\bibitem[{Greshake et~al.(2023)Greshake, Abdelnabi, Mishra, Endres, Holz, and
  Fritz}]{greshake2023youve}
Kai Greshake, Sahar Abdelnabi, Shailesh Mishra, Christoph Endres, Thorsten
  Holz, and Mario Fritz. 2023.
\newblock \href {http://arxiv.org/abs/2302.12173} {Not what you've signed up
  for: Compromising real-world llm-integrated applications with indirect prompt
  injection}.

\bibitem[{Guu et~al.(2020)Guu, Lee, Tung, Pasupat, and
  Chang}]{pmlr-v119-guu20a}
Kelvin Guu, Kenton Lee, Zora Tung, Panupong Pasupat, and Mingwei Chang. 2020.
\newblock \href {https://proceedings.mlr.press/v119/guu20a.html} {Retrieval
  augmented language model pre-training}.
\newblock In \emph{Proceedings of the 37th International Conference on Machine
  Learning}, volume 119 of \emph{Proceedings of Machine Learning Research},
  pages 3929--3938. PMLR.

\bibitem[{Hong et~al.(2020)Hong, Kang, Lim, and
  Myaeng}]{hong-etal-2020-handling}
Giwon Hong, Junmo Kang, Doyeon Lim, and Sung-Hyon Myaeng. 2020.
\newblock \href {https://doi.org/10.18653/v1/2020.coling-main.306} {Handling
  anomalies of synthetic questions in unsupervised question answering}.
\newblock In \emph{Proceedings of the 28th International Conference on
  Computational Linguistics}, pages 3441--3448, Barcelona, Spain (Online).
  International Committee on Computational Linguistics.

\bibitem[{Hong et~al.(2022)Hong, Kim, Kang, and Myaeng}]{hong-etal-2022-graph}
Giwon Hong, Jeonghwan Kim, Junmo Kang, and Sung-Hyon Myaeng. 2022.
\newblock \href {https://doi.org/10.18653/v1/2022.emnlp-main.702}
  {Graph-induced transformers for efficient multi-hop question answering}.
\newblock In \emph{Proceedings of the 2022 Conference on Empirical Methods in
  Natural Language Processing}, pages 10288--10294, Abu Dhabi, United Arab
  Emirates. Association for Computational Linguistics.

\bibitem[{Honnibal et~al.(2020)Honnibal, Montani, Van~Landeghem, Boyd
  et~al.}]{honnibal2020spacy}
Matthew Honnibal, Ines Montani, Sofie Van~Landeghem, Adriane Boyd, et~al. 2020.
\newblock spacy: Industrial-strength natural language processing in python.

\bibitem[{Izacard and Grave(2021)}]{izacard2021leveraging}
Gautier Izacard and {\'E}douard Grave. 2021.
\newblock Leveraging passage retrieval with generative models for open domain
  question answering.
\newblock In \emph{Proceedings of the 16th Conference of the European Chapter
  of the Association for Computational Linguistics: Main Volume}, pages
  874--880.

\bibitem[{Jang et~al.(2021)Jang, Kang, Hong, Myaeng, Park, Yoon, and
  Seo}]{jang-etal-2021-ultra}
Kyoung-Rok Jang, Junmo Kang, Giwon Hong, Sung-Hyon Myaeng, Joohee Park, Taewon
  Yoon, and Heecheol Seo. 2021.
\newblock \href {https://doi.org/10.18653/v1/2021.emnlp-main.78} {Ultra-high
  dimensional sparse representations with binarization for efficient text
  retrieval}.
\newblock In \emph{Proceedings of the 2021 Conference on Empirical Methods in
  Natural Language Processing}, pages 1016--1029, Online and Punta Cana,
  Dominican Republic. Association for Computational Linguistics.

\bibitem[{Joshi et~al.(2017)Joshi, Choi, Weld, and
  Zettlemoyer}]{joshi-etal-2017-triviaqa}
Mandar Joshi, Eunsol Choi, Daniel Weld, and Luke Zettlemoyer. 2017.
\newblock \href {https://doi.org/10.18653/v1/P17-1147} {{T}rivia{QA}: A large
  scale distantly supervised challenge dataset for reading comprehension}.
\newblock In \emph{Proceedings of the 55th Annual Meeting of the Association
  for Computational Linguistics (Volume 1: Long Papers)}, pages 1601--1611,
  Vancouver, Canada. Association for Computational Linguistics.

\bibitem[{Kamath et~al.(2020)Kamath, Jia, and Liang}]{kamath2020selective}
Amita Kamath, Robin Jia, and Percy Liang. 2020.
\newblock Selective question answering under domain shift.
\newblock In \emph{Proceedings of the 58th Annual Meeting of the Association
  for Computational Linguistics}, pages 5684--5696.

\bibitem[{Kang et~al.(2020)Kang, Hong, Puerto San~Roman, and
  Myaeng}]{kang-etal-2020-regularization}
Junmo Kang, Giwon Hong, Haritz Puerto San~Roman, and Sung-Hyon Myaeng. 2020.
\newblock \href {https://doi.org/10.18653/v1/2020.findings-emnlp.293}
  {Regularization of distinct strategies for unsupervised question generation}.
\newblock In \emph{Findings of the Association for Computational Linguistics:
  EMNLP 2020}, pages 3266--3277, Online. Association for Computational
  Linguistics.

\bibitem[{Karpukhin et~al.(2020)Karpukhin, Oguz, Min, Lewis, Wu, Edunov, Chen,
  and Yih}]{karpukhin2020dense}
Vladimir Karpukhin, Barlas Oguz, Sewon Min, Patrick Lewis, Ledell Wu, Sergey
  Edunov, Danqi Chen, and Wen-tau Yih. 2020.
\newblock Dense passage retrieval for open-domain question answering.
\newblock In \emph{Proceedings of the 2020 Conference on Empirical Methods in
  Natural Language Processing (EMNLP)}, pages 6769--6781.

\bibitem[{Kazemi et~al.(2023)Kazemi, Yuan, Bhatia, Kim, Xu, Imbrasaite, and
  Ramachandran}]{Kazemi2023BoardgameQAAD}
Mehran Kazemi, Quan Yuan, Deepti Bhatia, Najoung Kim, Xin Xu, Vaiva Imbrasaite,
  and Deepak Ramachandran. 2023.
\newblock Boardgameqa: A dataset for natural language reasoning with
  contradictory information.
\newblock \emph{ArXiv}, abs/2306.07934.

\bibitem[{Khalifa et~al.(2022)Khalifa, Logeswaran, Lee, Lee, and
  Wang}]{khalifa2022lepus}
Muhammad Khalifa, Lajanugen Logeswaran, Moontae Lee, Honglak Lee, and Lu~Wang.
  2022.
\newblock Lepus: Prompt-based unsupervised multi-hop reranking for open-domain
  qa.
\newblock \emph{arXiv preprint arXiv:2205.12650}.

\bibitem[{Kim et~al.(2021{\natexlab{a}})Kim, Hong, Kim, Kang, and
  Myaeng}]{kim-etal-2021-seen}
Jeonghwan Kim, Giwon Hong, Kyung-min Kim, Junmo Kang, and Sung-Hyon Myaeng.
  2021{\natexlab{a}}.
\newblock \href {https://doi.org/10.18653/v1/2021.emnlp-main.563} {Have you
  seen that number? investigating extrapolation in question answering models}.
\newblock In \emph{Proceedings of the 2021 Conference on Empirical Methods in
  Natural Language Processing}, pages 7031--7037, Online and Punta Cana,
  Dominican Republic. Association for Computational Linguistics.

\bibitem[{Kim et~al.(2023)Kim, Hong, Myaeng, and
  Whang}]{kim-etal-2023-fineprompt}
Jeonghwan Kim, Giwon Hong, Sung-Hyon Myaeng, and Joyce~Jiyoung Whang. 2023.
\newblock \href {https://doi.org/10.18653/v1/2023.findings-emnlp.245}
  {{F}ine{P}rompt: Unveiling the role of finetuned inductive bias on
  compositional reasoning in {GPT}-4}.
\newblock In \emph{Findings of the Association for Computational Linguistics:
  EMNLP 2023}, pages 3763--3775, Singapore. Association for Computational
  Linguistics.

\bibitem[{Kim et~al.(2022)Kim, Kang, Kim, Hong, and
  Myaeng}]{kim-etal-2022-exploiting}
Jeonghwan Kim, Junmo Kang, Kyung-min Kim, Giwon Hong, and Sung-Hyon Myaeng.
  2022.
\newblock \href {https://doi.org/10.18653/v1/2022.findings-naacl.138}
  {Exploiting numerical-contextual knowledge to improve numerical reasoning in
  question answering}.
\newblock In \emph{Findings of the Association for Computational Linguistics:
  NAACL 2022}, pages 1811--1821, Seattle, United States. Association for
  Computational Linguistics.

\bibitem[{Kim et~al.(2021{\natexlab{b}})Kim, Kang, Shin, and
  Myaeng}]{kim-etal-2021-distinguish}
Jeonghwan Kim, Junmo Kang, Suwon Shin, and Sung-Hyon Myaeng.
  2021{\natexlab{b}}.
\newblock \href {https://aclanthology.org/2021.hcinlp-1.9} {Can you distinguish
  truthful from fake reviews? user analysis and assistance tool for fake review
  detection}.
\newblock In \emph{Proceedings of the First Workshop on Bridging
  Human{--}Computer Interaction and Natural Language Processing}, pages 53--59,
  Online. Association for Computational Linguistics.

\bibitem[{Kwiatkowski et~al.(2019)Kwiatkowski, Palomaki, Redfield, Collins,
  Parikh, Alberti, Epstein, Polosukhin, Devlin, Lee, Toutanova, Jones, Kelcey,
  Chang, Dai, Uszkoreit, Le, and Petrov}]{kwiatkowski-etal-2019-natural}
Tom Kwiatkowski, Jennimaria Palomaki, Olivia Redfield, Michael Collins, Ankur
  Parikh, Chris Alberti, Danielle Epstein, Illia Polosukhin, Jacob Devlin,
  Kenton Lee, Kristina Toutanova, Llion Jones, Matthew Kelcey, Ming-Wei Chang,
  Andrew~M. Dai, Jakob Uszkoreit, Quoc Le, and Slav Petrov. 2019.
\newblock \href {https://doi.org/10.1162/tacl_a_00276} {Natural questions: A
  benchmark for question answering research}.
\newblock \emph{Transactions of the Association for Computational Linguistics},
  7:452--466.

\bibitem[{Lakhotia et~al.(2021)Lakhotia, Paranjape, Ghoshal, Yih, Mehdad, and
  Iyer}]{lakhotia-etal-2021-fid}
Kushal Lakhotia, Bhargavi Paranjape, Asish Ghoshal, Scott Yih, Yashar Mehdad,
  and Srini Iyer. 2021.
\newblock \href {https://doi.org/10.18653/v1/2021.emnlp-main.301} {{F}i{D}-ex:
  Improving sequence-to-sequence models for extractive rationale generation}.
\newblock In \emph{Proceedings of the 2021 Conference on Empirical Methods in
  Natural Language Processing}, pages 3712--3727, Online and Punta Cana,
  Dominican Republic. Association for Computational Linguistics.

\bibitem[{Le et~al.(2022)Le, Bai, and Ritter}]{Le2022FewShotAR}
Nghia Le, Fan Bai, and Alan Ritter. 2022.
\newblock Few-shot anaphora resolution in scientific protocols via mixtures of
  in-context experts.
\newblock In \emph{Findings of the Association for Computational Linguistics:
  EMNLP 2022}, Abu Dhabi, United Arab Emirates. Association for Computational
  Linguistics.

\bibitem[{Lewis et~al.(2020)Lewis, Perez, Piktus, Petroni, Karpukhin, Goyal,
  K{\"u}ttler, Lewis, Yih, Rockt{\"a}schel et~al.}]{lewis2020retrieval}
Patrick Lewis, Ethan Perez, Aleksandra Piktus, Fabio Petroni, Vladimir
  Karpukhin, Naman Goyal, Heinrich K{\"u}ttler, Mike Lewis, Wen-tau Yih, Tim
  Rockt{\"a}schel, et~al. 2020.
\newblock Retrieval-augmented generation for knowledge-intensive nlp tasks.
\newblock \emph{Advances in Neural Information Processing Systems},
  33:9459--9474.

\bibitem[{Lewis et~al.(2021)Lewis, Wu, Liu, Minervini, K{\"u}ttler, Piktus,
  Stenetorp, and Riedel}]{lewis-etal-2021-paq}
Patrick Lewis, Yuxiang Wu, Linqing Liu, Pasquale Minervini, Heinrich
  K{\"u}ttler, Aleksandra Piktus, Pontus Stenetorp, and Sebastian Riedel. 2021.
\newblock \href {https://doi.org/10.1162/tacl_a_00415} {{PAQ}: 65 million
  probably-asked questions and what you can do with them}.
\newblock \emph{Transactions of the Association for Computational Linguistics},
  9:1098--1115.

\bibitem[{Liu et~al.(2023)Liu, Yang, Jia, Zhang, Zhou, Dai, Yang, and
  Vosoughi}]{liu2023sociallyaligned}
Ruibo Liu, Ruixin Yang, Chenyan Jia, Ge~Zhang, Denny Zhou, Andrew~M. Dai, Diyi
  Yang, and Soroush Vosoughi. 2023.
\newblock \href {http://arxiv.org/abs/2305.16960} {Training socially aligned
  language models in simulated human society}.

\bibitem[{Longpre et~al.(2021)Longpre, Perisetla, Chen, Ramesh, DuBois, and
  Singh}]{longpre-etal-2021-entity}
Shayne Longpre, Kartik Perisetla, Anthony Chen, Nikhil Ramesh, Chris DuBois,
  and Sameer Singh. 2021.
\newblock \href {https://doi.org/10.18653/v1/2021.emnlp-main.565} {Entity-based
  knowledge conflicts in question answering}.
\newblock In \emph{Proceedings of the 2021 Conference on Empirical Methods in
  Natural Language Processing}, pages 7052--7063, Online and Punta Cana,
  Dominican Republic. Association for Computational Linguistics.

\bibitem[{Luo et~al.(2023)Luo, Chuang, Gong, Zhang, Kim, Wu, Fox, Meng, and
  Glass}]{luo2023sail}
Hongyin Luo, Yung-Sung Chuang, Yuan Gong, Tianhua Zhang, Yoon Kim, Xixin Wu,
  Danny Fox, Helen Meng, and James Glass. 2023.
\newblock \href {http://arxiv.org/abs/2305.15225} {Sail: Search-augmented
  instruction learning}.

\bibitem[{Min et~al.(2022)Min, Lewis, Hajishirzi, and
  Zettlemoyer}]{min-etal-2022-noisy}
Sewon Min, Mike Lewis, Hannaneh Hajishirzi, and Luke Zettlemoyer. 2022.
\newblock \href {https://doi.org/10.18653/v1/2022.acl-long.365} {Noisy channel
  language model prompting for few-shot text classification}.
\newblock In \emph{Proceedings of the 60th Annual Meeting of the Association
  for Computational Linguistics (Volume 1: Long Papers)}, pages 5316--5330,
  Dublin, Ireland. Association for Computational Linguistics.

\bibitem[{Min et~al.(2023)Min, Shi, Lewis, Chen, Yih, Hajishirzi, and
  Zettlemoyer}]{min-etal-2023-nonparametric}
Sewon Min, Weijia Shi, Mike Lewis, Xilun Chen, Wen-tau Yih, Hannaneh
  Hajishirzi, and Luke Zettlemoyer. 2023.
\newblock \href {https://doi.org/10.18653/v1/2023.findings-acl.132}
  {Nonparametric masked language modeling}.
\newblock In \emph{Findings of the Association for Computational Linguistics:
  ACL 2023}, pages 2097--2118, Toronto, Canada. Association for Computational
  Linguistics.

\bibitem[{Neeman et~al.(2023)Neeman, Aharoni, Honovich, Choshen, Szpektor, and
  Abend}]{neeman-etal-2023-disentqa}
Ella Neeman, Roee Aharoni, Or~Honovich, Leshem Choshen, Idan Szpektor, and Omri
  Abend. 2023.
\newblock \href {https://doi.org/10.18653/v1/2023.acl-long.559} {{D}isent{QA}:
  Disentangling parametric and contextual knowledge with counterfactual
  question answering}.
\newblock In \emph{Proceedings of the 61st Annual Meeting of the Association
  for Computational Linguistics (Volume 1: Long Papers)}, pages 10056--10070,
  Toronto, Canada. Association for Computational Linguistics.

\bibitem[{OpenAI(2023)}]{openai2023gpt4}
OpenAI. 2023.
\newblock \href {http://arxiv.org/abs/2303.08774} {Gpt-4 technical report}.

\bibitem[{Ouyang et~al.(2022)Ouyang, Wu, Jiang, Almeida, Wainwright, Mishkin,
  Zhang, Agarwal, Slama, Ray, Schulman, Hilton, Kelton, Miller, Simens, Askell,
  Welinder, Christiano, Leike, and Lowe}]{NEURIPS2022_b1efde53}
Long Ouyang, Jeffrey Wu, Xu~Jiang, Diogo Almeida, Carroll Wainwright, Pamela
  Mishkin, Chong Zhang, Sandhini Agarwal, Katarina Slama, Alex Ray, John
  Schulman, Jacob Hilton, Fraser Kelton, Luke Miller, Maddie Simens, Amanda
  Askell, Peter Welinder, Paul~F Christiano, Jan Leike, and Ryan Lowe. 2022.
\newblock \href
  {https://proceedings.neurips.cc/paper_files/paper/2022/file/b1efde53be364a73914f58805a001731-Paper-Conference.pdf}
  {Training language models to follow instructions with human feedback}.
\newblock In \emph{Advances in Neural Information Processing Systems},
  volume~35, pages 27730--27744. Curran Associates, Inc.

\bibitem[{Pan et~al.(2023)Pan, Pan, Chen, Nakov, Kan, and
  Wang}]{pan-etal-2023-risk}
Yikang Pan, Liangming Pan, Wenhu Chen, Preslav Nakov, Min-Yen Kan, and William
  Wang. 2023.
\newblock \href {https://doi.org/10.18653/v1/2023.findings-emnlp.97} {On the
  risk of misinformation pollution with large language models}.
\newblock In \emph{Findings of the Association for Computational Linguistics:
  EMNLP 2023}, pages 1389--1403, Singapore. Association for Computational
  Linguistics.

\bibitem[{Raffel et~al.(2020)Raffel, Shazeer, Roberts, Lee, Narang, Matena,
  Zhou, Li, and Liu}]{2020t5}
Colin Raffel, Noam Shazeer, Adam Roberts, Katherine Lee, Sharan Narang, Michael
  Matena, Yanqi Zhou, Wei Li, and Peter~J. Liu. 2020.
\newblock \href {http://jmlr.org/papers/v21/20-074.html} {Exploring the limits
  of transfer learning with a unified text-to-text transformer}.
\newblock \emph{Journal of Machine Learning Research}, 21(140):1--67.

\bibitem[{Robertson et~al.(2009)Robertson, Zaragoza
  et~al.}]{robertson2009probabilistic}
Stephen Robertson, Hugo Zaragoza, et~al. 2009.
\newblock The probabilistic relevance framework: Bm25 and beyond.
\newblock \emph{Foundations and Trends{\textregistered} in Information
  Retrieval}, 3(4):333--389.

\bibitem[{Si et~al.(2023)Si, Gan, Yang, Wang, Wang, Boyd-Graber, and
  Wang}]{si2023prompting}
Chenglei Si, Zhe Gan, Zhengyuan Yang, Shuohang Wang, Jianfeng Wang, Jordan~Lee
  Boyd-Graber, and Lijuan Wang. 2023.
\newblock \href {https://openreview.net/forum?id=98p5x51L5af} {Prompting
  {GPT}-3 to be reliable}.
\newblock In \emph{The Eleventh International Conference on Learning
  Representations}.

\bibitem[{Vig(2019)}]{vig-2019-multiscale}
Jesse Vig. 2019.
\newblock \href {https://doi.org/10.18653/v1/P19-3007} {A multiscale
  visualization of attention in the transformer model}.
\newblock In \emph{Proceedings of the 57th Annual Meeting of the Association
  for Computational Linguistics: System Demonstrations}, pages 37--42,
  Florence, Italy. Association for Computational Linguistics.

\bibitem[{Zhang et~al.(2021)Zhang, Gong, and Choi}]{zhang-etal-2021-knowing}
Shujian Zhang, Chengyue Gong, and Eunsol Choi. 2021.
\newblock \href {https://doi.org/10.18653/v1/2021.findings-acl.172} {Knowing
  more about questions can help: Improving calibration in question answering}.
\newblock In \emph{Findings of the Association for Computational Linguistics:
  ACL-IJCNLP 2021}, pages 1958--1970, Online. Association for Computational
  Linguistics.

\bibitem[{Zhao et~al.(2021)Zhao, Wallace, Feng, Klein, and
  Singh}]{pmlr-v139-zhao21c}
Zihao Zhao, Eric Wallace, Shi Feng, Dan Klein, and Sameer Singh. 2021.
\newblock \href {https://proceedings.mlr.press/v139/zhao21c.html} {Calibrate
  before use: Improving few-shot performance of language models}.
\newblock In \emph{Proceedings of the 38th International Conference on Machine
  Learning}, volume 139 of \emph{Proceedings of Machine Learning Research},
  pages 12697--12706. PMLR.

\bibitem[{Zhou et~al.(2023)Zhou, Zhang, Luo, Parker, and
  De~Choudhury}]{10.1145/3544548.3581318}
Jiawei Zhou, Yixuan Zhang, Qianni Luo, Andrea~G Parker, and Munmun
  De~Choudhury. 2023.
\newblock \href {https://doi.org/10.1145/3544548.3581318} {Synthetic lies:
  Understanding ai-generated misinformation and evaluating algorithmic and
  human solutions}.
\newblock In \emph{Proceedings of the 2023 CHI Conference on Human Factors in
  Computing Systems}, CHI '23, New York, NY, USA. Association for Computing
  Machinery.

\end{thebibliography}

\clearpage
\appendix

\section{Discussion}
\label{sec:faq}

\subsection{Why Combine GPT-3.5 and FiD?}
A crucial inquiry that may arise from our approach is \textit{Why do we need to combine GPT-3.5 and \texttt{\textbf{$\text{Disc}^{\text{FiD}}$}} despite its worse performance than the FiD counterpart?}
Note that our discriminator is easily trainable with our scalable perturbation framework (\S \ref{sec:gen_perturb}). In a low-resource setting, where downstream task instances are scarce, GPT-3.5's few-shot learning capability shines. The lightweight fine-tuned LMs trained on an easily accessible subtask (e.g., perturbation classification) can, therefore, maximize GPT-3.5's capability.

\subsection{On Perturbation Schemes}
Inconsistencies caused by conflicting information may occur for various reasons such as updated/outdated or fabricated/hallucinated information.
Our study explores \textit{the robustness of models under contradictory information, and the influence of varying degrees of noise}.
To facilitate a controllable study, we generate perturbed documents by adopting an entity-centric perturbation strategy \cite{longpre-etal-2021-entity}. While \citet{longpre-etal-2021-entity}'s entity perturbation framework has been widely adopted in several previous works \cite{chen2022rich, neeman-etal-2023-disentqa}, the perturbation framework faces a number of limitations as we elaborate in \S \ref{subsec:limitations_entity_framework}. Our work aims to overcome the confines of the entity-only perturbation framework and propose a new perturbation scheme using LLMs, with which we build \textsc{MacNoise}.

We measure the LMs' performance by explicitly controlling the proportion of perturbed documents (\textbf{0\%, 15\%, 25\%, 35\%}). The objective of this extensive study using the scalable and controllable framework is that the proportion of misleading information in the real-world is unknown, consistently changes, or varies depending on document sources. We believe that conflicts may potentially occur in other ways as well, but we clarify that exploring those is beyond the scope of our study.

\setlength{\tabcolsep}{2.3pt}
\begin{table}[t]
\small
\centering
\begin{tabular}{lccccccc} 
\textbf{Split} & \textbf{Total} & \textbf{N/A} & \textbf{PER} & \textbf{ORG} & \textbf{LOC} & \textbf{DATE} & \textbf{NUM}\\ 
\midrule
\multicolumn{8}{c}{\textbf{NQ-open}} \\
\midrule
Train & 79,168 & 27,916 & 20,136 & 2,611 & 4,311 & 5,343 & 3,628 \\
Dev & 8,757 & 3,099 & 2,872 & 394 & 461 & 1,365 & 566 \\
Test & 3,610 & 1,322 & 897 & 139 & 248 & 280 & 165 \\
\midrule
\multicolumn{8}{c}{\textbf{TQA-open}} \\
\midrule
Train & 78,785 & 40,252 & 19,107 & 5,838 & 10,141 & 1,264 & 2,183 \\
Dev & 8,837 & 4,542 & 2,120 & 665 & 1,123 & 163 & 224\\
Test & 11,313 & 2,891 & 4,162 & 2,683 & 1,017 & 142 & 418 \\
\bottomrule
\end{tabular}
\caption{NQ-Open and TQA-open dataset statistics and the type-wise count on the number of instances perturbed using a substitution framework in \citet{longpre-etal-2021-entity}. \textbf{N/A} denotes the instances with non-named entity answers that were not perturbed.}
\label{table:typewise_number}
\end{table}

\setlength{\tabcolsep}{2.5pt}
\begin{table}[t]
\small
\centering
\begin{tabular}{cccc}
\multicolumn{2}{c}{\textbf{Pert. \%}} & \multicolumn{2}{c}{\textbf{\# of Pert. Documents}} \\ 
\midrule
& & \textbf{NQ-Open} & \textbf{TQA-Open}  \\
\midrule
\multirow{3}{*}{\shortstack{\textbf{Dev}}}
& \textbf{30\%} & 191 (14.92\%) & 199 (15.54\%) \\
& \textbf{50\%} & 312 (24.38\%) & 317 (24.76\%)  \\
& \textbf{75\%} & 453 (35.39\%) & 453 (35.39\%) \\
\midrule
\multirow{3}{*}{\shortstack{\textbf{Test}}}
& \textbf{30\%} & 1,369 (14.96\%) & 3,356 (15.03\%) \\
& \textbf{50\%} & 2,308 (25.22\%) & 5,572 (24.96\%) \\
& \textbf{75\%} & 3,471 (37.93\%) & 7,811 (34.99\%) \\
\bottomrule
\end{tabular}
\caption{Statistic on the number of documents perturbed from the 256 dev instances and full test instances sampled for our evaluation. Each row represents the perturbation probability, and the \textbf{\# of Documents} refer to the perturbed documents and their portion in the percentage out of the 1,280 documents for dev (from which 48.05\% were perturbable) and the 9,150 documents for test (from which 50.67\% were perturbable) in case of NQ-open. For TQA-open, the percentage is calculated based on the 1,280 documents for dev (from which 67.57\% were perturbable) and the 14,974 documents for test (from which 67.08\% were perturbable).}
\label{table:pert_ratio_stat}
\end{table}

\begin{table*}[h!]
\label{table:prompts}
\begin{center}
\begin{adjustbox}{width=0.8\textwidth}
\begin{tabular}{m{12cm}}
\toprule
\textsc{MacNoise} Prompt \\
\midrule
\tiny{\texttt{You are a novel writing AI. Your job is to make up a story based on the following information.\newline
You will be given a question (preceded by "Question:"), a document (preceded by "Document:") and \newline
the corresponding answer ("Answer:"), and you will be asked to create a novel story after ("Revised Document:").
Note, there can be multiple answers (['answer1', 'answer2', ...]) to a given question and document pair.
\newline
Now, you should creatively rewrite the document so that the document has a different answer than the given answer(s).\newline
\newline
The rewritten document must adhere to all of the following rules:\newline
1) The rewritten document must be answerable by the question. \newline
The information (e.g., entities, phrases) explicitly in the question should not be changed from the original document.\newline
2) The rewritten document should be similar in length to the given original document above.\newline
3) The rewritten document should not contain the original answer. \newline
If the original answer cannot be removed from the document, rewrite the document so the semantics negate / do not support the answer.\newline
\newline
The following are the possible rewriting strategies:\newline
1) Rewrite the document so the passage no longer supports the answer.\newline
2) Replace the entity in the passage.\newline
3) Negate the sentence the answer span exists so that the original answer span is no longer the answer.
\newline
Make sure that the rewritten document is in a completely different style than the original document, and correctly generate punctuations like periods (".") and commas (","). \newline
\newline
You must give your rewritten document only after "Revised Document:".
}} \\


\bottomrule
\end{tabular}
\end{adjustbox}
\caption{A prompt used for counterfactual document generation using large language models.}
\label{table:macnoise_prompt}
\end{center}
\end{table*}

\section{Generation of Counterfactual Documents to Infuse Conflicting Information}
\label{sec:gen_perturb_detail}

Our work mainly focuses on improving the retrieval-augmented LMs for ODQA when presented with a mixed bag of gold and counterfactual documents. 

\subsection{Entity Perturbation \cite{longpre-etal-2021-entity}}
\label{subsec:generate_longpre}
The counterfactual documents are generated with the entity-perturbation framework proposed in \citet{longpre-etal-2021-entity}\footnote{https://github.com/apple/ml-knowledge-conflicts/tree/main released under \texttt{Copyright (C) 2021 Apple Inc. All Rights Reserved.}}; we use the \texttt{corpus-substitution} scheme in this work. While the previous works \cite{chen2022rich, neeman-etal-2023-disentqa, si2023prompting} use the framework to investigate the effect of knowledge memorization, our work leverages the entity substitution to generate counterfactual documents that contradict what the model has already learned.

We first identify the instances that have the five named entities - PER, ORG, LOC, DATE, and NUM - as their gold answer (as defined in the previous work), and tag each gold named entity answer with a Named-Entity Recognition (NER) tool\footnote{We used the spaCy NER tool (version 3.5.1) \cite{honnibal2020spacy}, an open-sourced natural language processing tool, released under \texttt{The MIT License (MIT).}}. Here, we define the "perturbable" documents as those that contain one of the five NER-typed entities as their answers, or non-perturbable otherwise.
Then, we use a set of retrieved documents using DPR \cite{karpukhin2020dense}, which was provided in the official repository of FiD \cite{izacard2021leveraging}\footnote{https://github.com/facebookresearch/FiD, released under the \texttt{Attribution-NonCommercial 4.0 International} license.} and find the spans in the documents that overlap with the gold answers. We then perturb each document with certain probabilities by substituting every named entity answer with a randomly sampled named entity. To avoid shortcuts and make the perturbed document discrimination task more challenging, we sample from a pool of entities of the same type as the substituted entity, e.g., Michael Jordan (PER) is replaced with Kobe Bryant (PER).
Table \ref{table:typewise_number} shows an overview of the NQ-Open dataset \cite{kwiatkowski-etal-2019-natural} and TQA-Open dataset \cite{joshi-etal-2017-triviaqa} used in this work and the type-wise number of instances that have named entities as their answers.

To give a detailed overview on the change in the number of documents with the increasing perturbation probability (\textbf{Pert. \%}), we also provide the perturbed document statistic in Table \ref{table:pert_ratio_stat}. The statistic elaborates the details about the perturbable documents within the sampled 256 dev set instances and the full test set instances from the NQ-Open dataset in \S \ref{sec:eval_longpre}. The "full" test set in our work refers to the 1,830 instances from the NQ-Open test set and 4,464 for the TQA-open test set; these are the instances that contain (i) perturbable passages which (ii) lie within the top 5 passages scored by DPR. In generating our training dataset, based on NQ-open, using \citet{longpre-etal-2021-entity}, we apply the same aforementioned entity perturbation strategy. For the training details, refer to Appendix \ref{subsec:settings_fid}

\subsection{\textsc{MacNoise}: Machine-Generated Perturbation}
\label{subsec:generate_macnoise}
Our \textsc{MacNoise} dataset also follows the same statistic as the dataset described in the previous Appendix section (\S \ref{subsec:generate_longpre}); since only answer-containing documents can be perturbed, meaning both \textsc{MacNoise} and entity perturbation were applied to the same subset, the statistics are identical to each other. The difference is that the perturbed documents for \textsc{MacNoise} were generated by
\begin{itemize}
    \item \texttt{GPT-3.5-turbo}: Used to generate the training dataset using NQ-open. This training dataset is part of \textsc{MacNoise}.
    \item GPT-4: Used to generate the evaluation dataset for NQ-open and TQA-open. This evaluation dataset is part of \textsc{MacNoise}.
\end{itemize}
The instruction prompt template used to generate the perturbed documents in \textsc{MacNoise} is presented in Table \ref{table:macnoise_prompt}. To deal with the extensive cost of generating all the perturbable training documents, we truncate the number of documents perturbed to 20 ($T=20$). In the case of building \textsc{MacNoise} dataset for TQA-open, we added three quality examples to the prompt as in-context demonstrations, where the examples are sampled from the earlier established dataset for NQ-open. This allowed us to ensure consistency in data quality, addressing the issue of OpenAI models that are subject to change over time, which we empirically encountered after the NQ-open creation with \textsc{MacNoise}.

\section{Details of Experimental Settings}
\label{sec:fid_train_eval}

\subsection{Overview}
\label{subsec:exp_datasets}

\paragraph{Models}
In this section, we provide the list of models we used in this work as our baseline for ODQA:
\begin{itemize}
    \item FiD \cite{izacard2021leveraging}: The retrieval-augmented LM used in our experiment. This includes our (i) \textbf{FiD (Semi-Parametric w/ \texttt{\textbf{$\text{Disc}^{\text{FiD}}$}})} setting in which we fine-tune the discriminator with either the entity-perturbed NQ-open or \textsc{MacNoise}, and the Semi-Parametric setting.
    \item GPT-3.5 \cite{NEURIPS2020_1457c0d6}: The LLM used in our experiment; the GPT-3.5 model we use as our baseline is \texttt{text-davinci-003}. We use the prompts in Figure \ref{fig:prompt_designs} for evaluation. This includes the \textbf{GPT-3.5 (Semi-Parametric w/ \texttt{\textbf{$\text{Disc}^{\text{Inst}}$}})} setting.
\end{itemize}

\paragraph{Datasets}
The datasets we based our perturbation schemes on are as follows:

\begin{itemize}
    \item Natural Questions (NQ) \cite{kwiatkowski-etal-2019-natural}\footnote{The dataset is released under the \texttt{Creative Commons Share-Alike 3.0} license.}: NQ is an English QA dataset consisting of real queries submitted to the Google search engine and Wikipedia documents. We used the open version of the NQ dataset (NQ-open) along with a set of documents retrieved using DPR \cite{karpukhin2020dense}. Due to the API budget constraint, we sample 256 dev set as in \citet{Le2022FewShotAR}, while using a full test set of 1,830 instances. There are a total of 79,168 training instances, 8,757 dev instances (from which 256 are sampled due to API budget constraint), and 3,610 test instances (from which 1,830 instances were perturbable). We provide additional details about generating the training and evaluation dataset in Appendices \ref{subsec:generate_longpre} and \ref{subsec:generate_macnoise}.
    
    \item TriviaQA (TQA) \cite{joshi-etal-2017-triviaqa}\footnote{The dataset is released under the \texttt{Apache 2.0} license.}: TQA is another English-oriented QA dataset, featuring queries sourced from a collection of 14 trivia and quiz-league websites. Specifically, we used the open, unfiltered version of TQA-open, akin to the process with NQ-open, retrieving documents from Wikipedia using DPR as in FiD \cite{lakhotia-etal-2021-fid}. Due to the API budget constraint, we sample 256 instances from the dev set, which consists of a total of 8,837.
\end{itemize}

\paragraph{Perturbation Schemes}
In this section, we provide a list of the entity perturbation schemes we used to perturb the datasets:
\begin{itemize}
    \item Entity Perturbation \cite{longpre-etal-2021-entity}: This method involves the direct replacement of one target entity with another random entity of the same type. The details of generating this dataset are provided in Appendix \ref{subsec:generate_longpre}.
    \item \textsc{MacNoise}: Our new machine-generated noise dataset created by \texttt{GPT-3.5-turbo} (for training dataset) and GPT-4 (for evaluation dataset) using the prompt given in Table \ref{table:macnoise_prompt}. For additional details, refer to Appendix \ref{subsec:generate_macnoise}. 
\end{itemize}

\subsection{Settings of FiD-based Models}
\label{subsec:settings_fid}
FiD\footnote{The models were trained using GeForce RTX3090 (24GB VRAM), AMD Ryzen Threadripper 3960X, and 128GiB RAM. The model training took approximately 80 hours.}
used in this work was based on T5-base (220M parameters) and trained to use a fewer number of retrieved passages due to our computing resource constraints. While FiD's base setting uses $T=100$ retrieved passages to answer open-domain questions, our work only considers $T=50$ for the entity perturbation scheme \cite{longpre-etal-2021-entity} and $T=20$ for \textsc{MacNoise}. This, however, does not present an issue to our study, since the findings in \citet{chen2022rich} show that FiD tends to focus its attention on the top $N$, where $N \leq 20$, retrieved documents when generating an answer. For model training, the perturbable probabilities were set to 30\%, 50\%, and 75\% to match the dev/test sets perturbed portions 15\%, 25\%, 35\%, respectively (Table \ref{table:pert_ratio_stat}). During training, every document undergoes random perturbation based on set probabilities, unlike during evaluation where perturbations are pre-defined. 
Our model was fine-tuned in the above settings independently.
This fine-tuned model was used in our experiments throughout. We believe that this setting is valid because in real life, we can sample from the real-world Web and identify the sampled distribution of misleading information.

\setlength{\tabcolsep}{3pt}
\begin{table}[t!]
\small
\begin{center}
\begin{tabular}{lc}
    \toprule
    \textbf{Hyperparameter} & \textbf{FiD}   \\
    \midrule\midrule
    Batch size   &   1 \\
    Gradient Accumulation & 64 \\
    Hidden size   &   768 \\
    Max. Sequence length   &  200 \\
    Learning rate   &   1e-4 \\
    Optimizer   &   AdamW \\
    Seed   &   42  \\
    \bottomrule
\end{tabular}
\end{center}
\caption{Hyperparameters of the fine-tuned FiD in this work. We set the gradient accumulation to 64 to account for the batch size in the original FiD \cite{izacard2021leveraging}.
Each passage is of Max. Sequence Length.}
\label{table:hyperparameter}
\end{table}

We also provide the important hyperparameters used to train our \textbf{FiD (Semi-Parametric w/ \texttt{\textbf{$\text{Disc}^{\text{FiD}}$}})} model (Table \ref{table:hyperparameter}). For all the other settings, including the size of the training dataset and the gradient steps, we follow the settings specified in the original FiD. Since our experiments demonstrated clear results sufficiently to validate our hypothesis made in this work, we did not perform a hyperparameter search, and the models were trained once with a fixed seed.

\subsection{Settings of Large Language Models}
\label{subsec:settings_gpt}

\setlength{\tabcolsep}{3pt}
\begin{table}[t!]
\small
\begin{center}
\begin{tabular}{lccc}
    \toprule
    \textbf{Hyperparameter} & \textbf{GPT-3.5}   &
    \textbf{ChatGPT}  &
    \textbf{GPT-4}  \\
    \midrule\midrule
    Context length   & 4,097 & 4,097 & 8,192 \\
    top\_p & 1.0 & 1.0 & 1.0  \\
    temperature  & 0.0 & 0.0 & 0.0  \\
    logprobs & 10 & N/A & N/A\\
    \bottomrule
\end{tabular}
\end{center}
\caption{Hyperparemters of the GPT-3.5 (\texttt{text-davinci-003}), ChatGPT (\texttt{GPT-3.5-turbo}), and GPT-4 used in our experiments.}
\label{table:hyperparameter_gpt}
\end{table}

Large Language Models (LLMs) used in this work are twofold: our baseline for ODQA\footnote{A total cost of approximately \$5,500 was incurred for API usage for ODQA experiments.} (text-davinci-003) and perturbation sources for our \textsc{MacNoise} dataset (\texttt{GPT-3.5-turbo} and GPT-4). We use the aforementioned LLMs through black-box API calls, and we provide the hyperparameters we used in API requests in Table \ref{table:hyperparameter_gpt}. We set \texttt{logprobs} as 10 for GPT-3.5 (\texttt{GPT-3.5-turbo}) to get top-10 generated answers for the ensemble strategy described in Appendix \ref{sec:ensemble_results}. For prompt designs, refer to Appendices \ref{subsec:generate_macnoise} (ODQA baseline) and \ref{sec:prompt_design} (generating \textsc{MacNoise} dataset). We set the number of documents used during evaluation to $T=5$, since the context window of GPT-3.5 is limited.

\subsection{Joint Training on \textsc{MacNoise} and \citet{longpre-etal-2021-entity}}
As discussed in \S \ref{subsec:complementarity}, our work further investigates the transferability and complementarity of the entity-perturbed and LLM-perturbed datasets in an effort to address both perturbation schemes with our fine-tuned discriminator model. We jointly fine-tune the \textbf{FiD (Semi-Parametric w/ \texttt{\textbf{$\text{Disc}^{\text{FiD}}$}})} model with both the entity-perturbed and LLM-perturbed NQ-open datasets by simply aggregating the two datasets together to form a joint training dataset as a whole. Here, we use the same number of documents ($T=20$) as the models for \textsc{MacNoise} do to make the resulting data balanced in terms of the perturbation type.

\subsection{LLM Prompt Designs for ODQA}
\label{sec:prompt_design}

\begin{figure*}[tph!]
    \centering
    \includegraphics[scale=0.47]{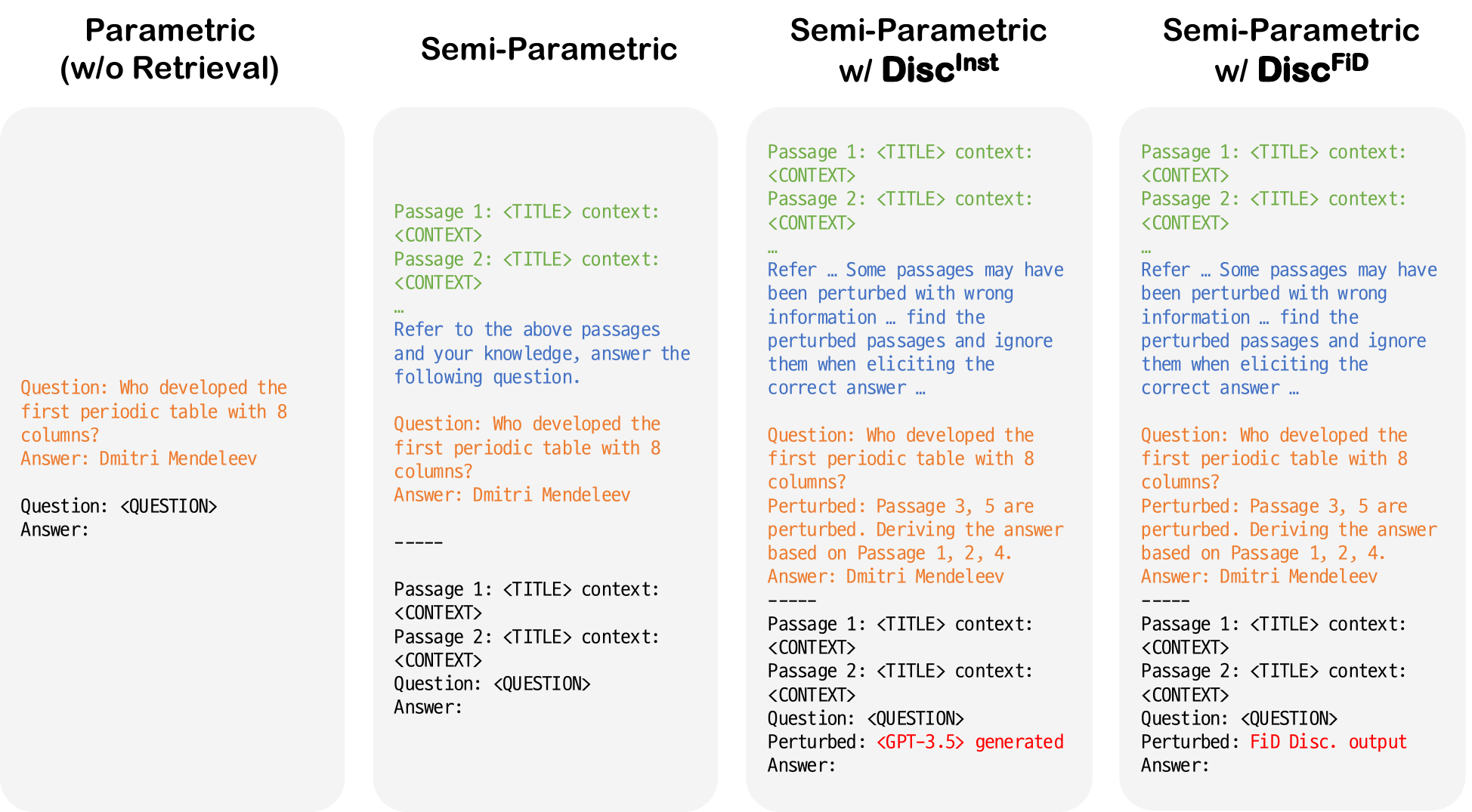}
    \caption{The prompt variants used in our experiments to evaluate the robustness of GPT-3.5 \texttt{text-davinci-003} when given a mixed bag of conflicting information-infused documents. The text in \textcolor{orange}{\textbf{orange}} refers to the in-context QA sample from the training data, \textcolor{green}{\textbf{green}} refers to the in-context retrieved document that corresponds to the \textcolor{orange}{\textbf{QA pair}}, \textcolor{blue}{\textbf{blue}} refers to the prompt, and \textcolor{red}{\textbf{red}} refers to either the GPT-3.5 generated perturbation classification, or the output of the FiD's discriminator fed straight into the prompt. The \textcolor{green}{\textbf{in-context sample documents}} may or may not be perturbed. The text in black refers to the evaluation instance.}
    \label{fig:prompt_designs}
\end{figure*}

In Figure \ref{fig:prompt_designs}, we explain in detail the design of our prompts used for ODQA. We divide the prompts into four discrete categories, with each one representing one of the four model settings in \S \ref{sec:eval_longpre}. Following the findings in \citet{khalifa2022lepus}, we place the instruction prompt ("\textit{Refer to the above documents and your knowledge ...}") after the retrieved documents, which takes the advantage of the \textit{recency bias} phenomenon evidenced in a previous work \cite{pmlr-v139-zhao21c}. 
The retrieved documents that precede the instruction are from the $k$ in-context instances ($k=5$) sampled from the held-out set; the held-out set refers to the remaining dev set instances aside from the 256 randomly sampled dev set used in our experiments.
To maximize the effect of our ensembling strategy, we sample the $k$ instances so each has a unique answer NER type and a different number of perturbed documents.
We then ensemble \cite{min-etal-2022-noisy, Le2022FewShotAR} over $k$ separate one-shot iterations for a single test instance to mitigate the in-context sample sensitivity observed in \citet{pmlr-v139-zhao21c}. Our approach ensembles over the $k$ iterations by marginalizing over the probability of the top 10 generated answers and chooses an answer with the maximum probability (Refer to Appendix \ref{sec:ensemble_results}).
Following the prompt is the one-shot in-context QA pair that guides the model to generate an appropriate answer given a set of retrieved documents and a question. The \textbf{Perturbed:} prompt that follows the question and GPT-3.5's generated response enables the model to discern perturbed from original documents. Note that what comes after the \textbf{Perturbed:} can be explicitly replaced with the FiD's jointly trained \texttt{\textbf{$\text{Disc}^{\text{FiD}}$}} output.

\section{Additional Experimental Results}

\subsection{Additional Results on Entity Perturbation}

\setlength{\tabcolsep}{2.5pt}
\begin{table}[t]
\small
\centering
\begin{tabular}{lccccccc} 
\toprule
\multirow{2}{*}{} & \multicolumn{3}{c}{\textbf{FiD}} & \multicolumn{4}{c}{\textbf{GPT-3.5}}\\ 
\cmidrule(lr){2-4}\cmidrule(lr){5-8}
& \textbf{Prec.} & \textbf{Rec.} & \textbf{F1} & & \textbf{Prec.} & \textbf{Rec.} & \textbf{F1} \\
\midrule
 \textbf{15\%} & 93.60 & 61.26 & 74.05 &  & 20.14 & 49.11 & 22.67 \\
 \textbf{25\%} & 98.51 & 63.78 & 77.43 &  & 30.29 & 48.59 & 37.32 \\
 \textbf{35\%} & 96.28 & 68.65 & 80.15 &  & 42.03 & 49.14 & 45.31 \\
\bottomrule
\end{tabular}
\caption{Classification performance of our discriminator on the sampled entity-perturbed NQ-open set (256 instances). Each row corresponds to perturbation \%. 
}
\label{table:disc_performance_dev}
\end{table}

\setlength{\tabcolsep}{2.5pt}
\begin{table}[t!]
\small
\centering
\begin{tabular}{lccccccc} 
\toprule
\multirow{2}{*}{} & \multicolumn{3}{c}{\textbf{FiD}} & \multicolumn{4}{c}{\textbf{GPT-3.5}}\\ 
\cmidrule(lr){2-4}\cmidrule(lr){5-8}
& \textbf{Prec.} & \textbf{Rec.} & \textbf{F1} & & \textbf{Prec.} & \textbf{Rec.} & \textbf{F1} \\
\midrule
 \textbf{15\%} & 80.42 & 57.79 & 67.25 &  & 16.94 & 48.34 & 25.09 \\
 \textbf{25\%} & 90.20 & 58.04 & 70.63 &  & 24.85 & 48.08 & 32.76 \\
 \textbf{35\%} & 94.35 & 58.94 & 72.55 &  & 39.89 & 51.21 & 44.85 \\
\bottomrule
\end{tabular}
\caption{Classification performance of our discriminator on the sampled entity-perturbed TQA-open dev set.}
\label{table:disc_tqa_longpre_dev}
\end{table}

\label{subsec:addl_result_entpert}
\paragraph{Classification}
In Table \ref{table:disc_performance_dev} and Table \ref{table:disc_tqa_longpre_dev}, we provide the performance of our discriminator on our sampled NQ-Open and TQA-open dev sets, respectively. The FiD result shows that the discriminator classifies perturbed and original documents with high precision, while recall lags behind.

\begin{figure}[th!]
    \centering
    \includegraphics[scale=0.26]{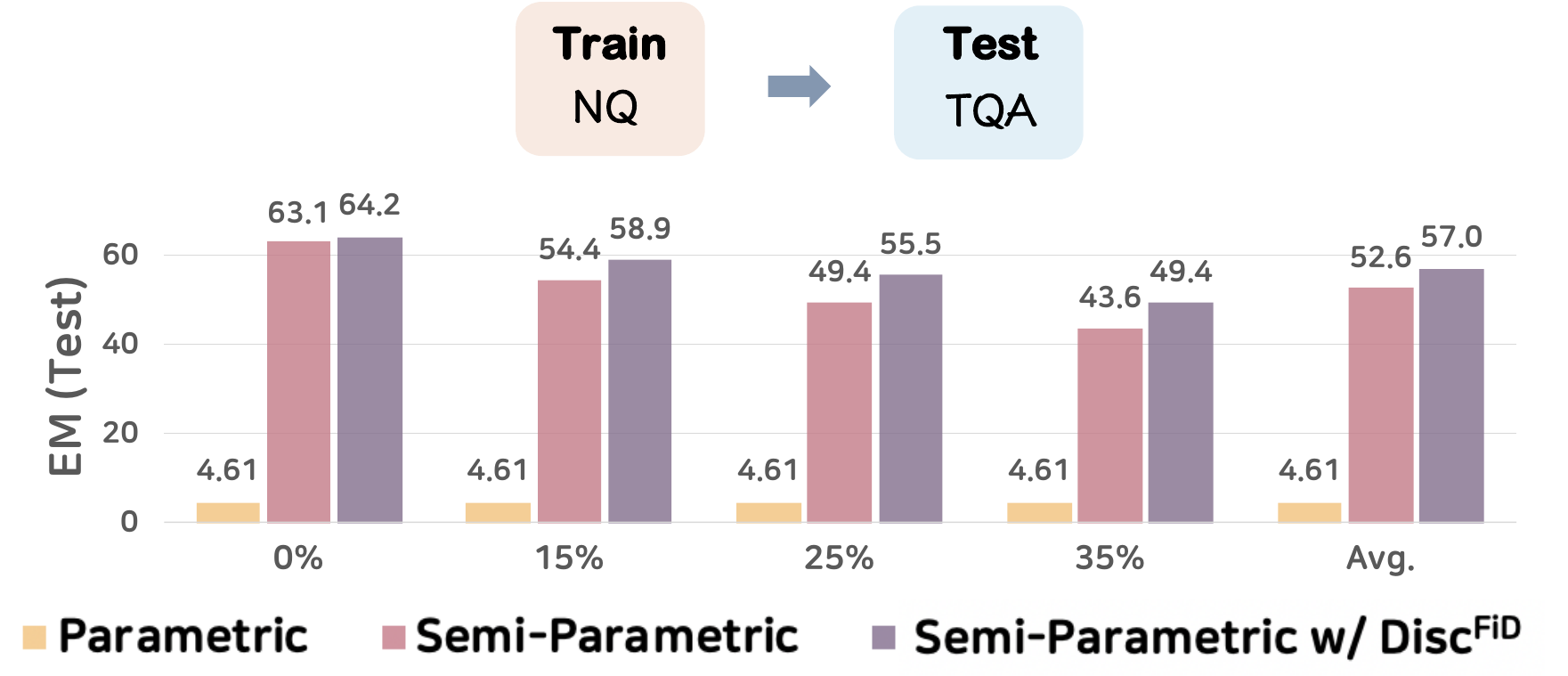}
    \caption{Results of FiD-based models on TQA-open (\textbf{test}). Models are trained on NQ-open and evaluated on TQA-open to examine the transferability of the robustness acquired through our method.} 
    \label{fig:triviaqa_test}
\end{figure}

\paragraph{Transferability to TQA-open}
We also demonstrate in Figure \ref{fig:triviaqa_test} the transferability of our NQ-open fine-tuned \textbf{FiD (Semi-Parametric w/ \texttt{\textbf{$\text{Disc}^{\text{FiD}}$}})} to TQA-open \textbf{test} dataset. The results demonstrate similar trends as those shown in \S \ref{sec:task_transfer_tqa}.

\subsection{Additional Results on \textsc{MacNoise}}
\label{subsec:addl_result_macnoise}

\paragraph{Classification}
\setlength{\tabcolsep}{2.5pt}
\begin{table}[t!]
\small
\centering
\begin{tabular}{lccccccc} 
\toprule
\multirow{2}{*}{} & \multicolumn{3}{c}{\textbf{FiD}} & \multicolumn{4}{c}{\textbf{GPT-3.5}}\\ 
\cmidrule(lr){2-4}\cmidrule(lr){5-8}
& \textbf{Prec.} & \textbf{Rec.} & \textbf{F1} & & \textbf{Prec.} & \textbf{Rec.} & \textbf{F1} \\
\midrule
 \textbf{15\%} & 94.32 & 41.71 & 57.84 &  & 14.89 & 51.46 & 23.09 \\
 \textbf{25\%} & 93.98 & 49.21 & 64.60 &  & 23.36 & 52.24 & 32.28 \\
 \textbf{35\%} & 93.77 & 53.20 & 67.89 &  & 36.46 & 54.30 & 43.63 \\
\bottomrule
\end{tabular}
\caption{Classification performance of our discriminator on the sampled TQA-open dev set with \textsc{MacNoise}.}
\label{table:disc_tqa_macnosie_dev}
\end{table}
In Table \ref{table:disc_tqa_macnosie_dev}, we provide the performance of our discriminator on our \textsc{MacNoise} TQA-open dev set. The FiD result shows that the discriminator classifies perturbed and original documents with high precision, while recall lags behind.

\paragraph{Enhanced Stability in GPT-3.5}

\begin{figure}[t!]
    \centering
    \includegraphics[scale=0.36]{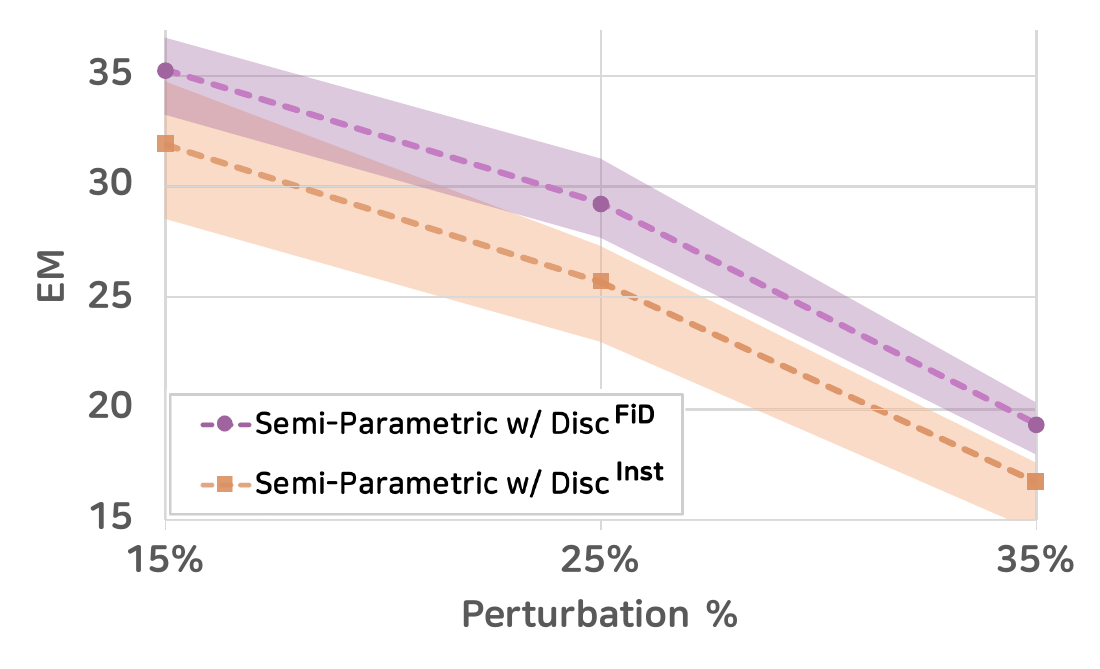}
    \caption{Comparison of GPT-3.5's stability for each discriminator setting on \textsc{MacNoise}.
    The shaded area represents the variance computed between the best and worst EM.} \label{fig:stability_figure_macnoise}
\end{figure}

Injecting the output decisions of our fine-tuned discriminator on \textsc{MacNoise} into GPT-3.5's prompts, as shown in Figure \ref{fig:stability_figure_macnoise}, notably improves the stability of the LLM prediction. Similar to the result in \S \ref{subsec:enhanced_icl_stability}, the dotted lines in Figure \ref{fig:stability_figure_macnoise} represent the average values over the ensemble, and the top and bottom of the shaded regions represent the worst and best cases, respectively.

\subsection{Qualitative Analysis on the Cross-Attention Weights of FiD models} 
\begin{figure*}[th!]
    \centering
    \includegraphics[scale=0.465]{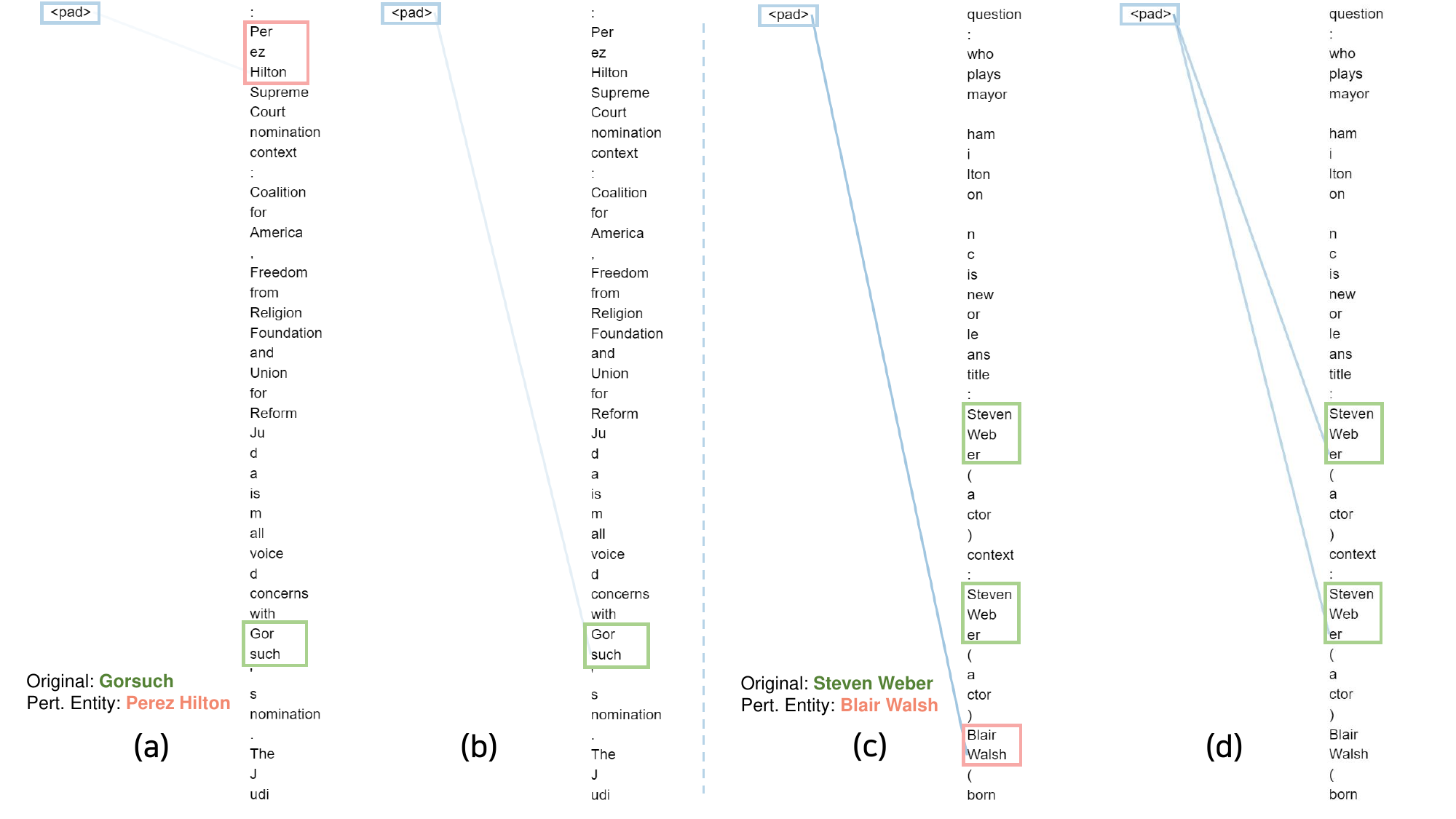}
    \caption{Illustration of our qualitative case study on the cross-attention weights. \texttt{(a)}, \texttt{(b)} and \texttt{(c)}, \texttt{(d)} are the before (\textbf{FiD (Semi-Parametric)}) and after (\textbf{FiD (Semi-Parametric w/ \texttt{\textbf{$\text{Disc}^{\text{FiD}}$}})}) screenshots of our attention scores, respectively. The perturbed entities are represented in \textcolor{red}{\textbf{red}} and the original entities are represented in \textcolor{green}{\textbf{green}}. The \texttt{<pad>} tokens are the starting input token to the FiD decoder and the lines denote the decoder's last layer cross attention to the encoder's output representations (represented here as input sequences).}
    \label{fig:case_study}
\end{figure*}

To investigate the effect of the learned discriminator on the answer generation by distinguishing perturbed from original entities, we conduct a qualitative study on the cross attentions of the samples shown in Figure \ref{fig:case_study}. The blue lines visualized\footnote{The weights were visualized using BertViz \cite{vig-2019-multiscale}.} denote the attention weight from the last layer of the decoder (i.e., starting token) to the encoder's output representations (i.e., input documents). In the first case, \texttt{(a)} shows that given a counterfactual entity, \texttt{Perez Hilton}, the \textbf{FiD (Semi-Parametric)} setting does not prevent the decoder from attending to the perturbed entity, neglecting the original entity, \texttt{Gorsuch}. On the contrary, in \texttt{(b)}, our \textbf{FiD (Semi-Parametric w/ \texttt{\textbf{$\text{Disc}^{\text{FiD}}$}})} setting, the decoder successfully attends to the original entity, \texttt{Gorsuch}, even in the presence of the perturbed entity. We also provide an additional before and after case in \texttt{(c)} and \texttt{(d)}, where the original entity, \texttt{Steven Weber} is replaced by \texttt{Blair Walsh}. In \texttt{(c)}, we show that \textbf{FiD (Semi-Parametric)} strongly attends to \texttt{Blair Walsh}, the perturbed entity, even in the presence of the two original entity spans in the given context. With our discriminator, we show in \texttt{(d)} that the model now attends to the two original entity spans correctly, successfully neglecting the perturbed entity. These cases serve as a testament that our learned discriminator enables the model to effectively control its attention from context-irrelevant, counterfactual entity to the original entity.

\subsection{Ensemble Strategy in GPT-3.5}
\label{sec:ensemble_results}

 \begin{table}[t!]
\small
\tabcolsep=0.1cm\centering
\begin{tabular}[t]{ lcc } 
\toprule
     &
    \textbf{Ensemble} &
    \textbf{Average} \\
\midrule
  \textbf{0\%} & \textbf{51.17} & 49.14  \\ 
  \textbf{15\%} & \textbf{42.18} & 40.70 \\ 
  \textbf{25\%} & \textbf{33.98} & 33.90\\
  \textbf{35\%} & \textbf{27.34} & 25.78  \\
\bottomrule
\end{tabular}
\caption{Comparison between the ensemble of the top-10 probabilities of the generated answer over $k=5$ iterations and an average of output scores over the iterations for Semi-Parametric w/ \texttt{\textbf{$\text{Disc}^{\text{FiD}}$}}.}
\label{table:ensemble_result}
\end{table}

\begin{table*}[ht!]
\centering
\setlength{\tabcolsep}{10pt}
\begin{tabular}{llccccc}
\toprule
\multirow{2.5}{*}{\shortstack[l]{\textbf{Number of}\\\textbf{Samples}}} & \multirow{2.5}{*}{\textbf{Method}} & \multicolumn{5}{c}{\textbf{Perturbation \% (Dev)}} \\
\cmidrule(lr){3-7}
 & & \textbf{0\%} & \textbf{15\%} & \textbf{25\%} & \textbf{35\%} & \textbf{Avg.} \\ 
\midrule

\multirow{5}{*}{\shortstack{$k=1$}} 
 & Parametric (w/o Retrieval) & \multicolumn{4}{c}{33.66} & 33.66 \\
 & Semi-Parametric & 50.22 & 42.95 & 35.25 & 22.24 & 37.67 \\
 & Semi-Parametric w/ \texttt{\textbf{$\text{Disc}^{\text{Inst}}$}} & 52.73 & 45.96 & 38.91 & 26.12 & 40.93 \\
 & Semi-parametric w/ \texttt{\textbf{$\text{Disc}^{\text{FiD}}$}} & \textbf{53.88} & \textbf{46.94} & \textbf{40.82} & \textbf{27.54} & \textbf{42.30} \\
 \midrule
\multirow{5}{*}{\shortstack{$k=2$}} 
 & Parametric (w/o Retrieval) & \multicolumn{4}{c}{35.90} & 35.90 \\
 & Semi-Parametric & 51.97 & 44.10 & 36.99 & 23.33 & 39.10 \\
 & Semi-Parametric w/ \texttt{\textbf{$\text{Disc}^{\text{Inst}}$}} & 54.48 & 47.05 & 39.78 & 27.10 & 42.10 \\
 & Semi-parametric w/ \texttt{\textbf{$\text{Disc}^{\text{FiD}}$}} & \textbf{55.36} & \textbf{48.85} & \textbf{42.35} & \textbf{28.52} & \textbf{43.77} \\
\midrule
\multirow{5}{*}{\shortstack{$k=3$}} 
 & Parametric (w/o Retrieval) & \multicolumn{4}{c}{36.50} & 36.50 \\
 & Semi-Parametric & 52.40 & 45.25 & 38.09 & 24.43 & 40.04 \\
 & Semi-Parametric w/ \texttt{\textbf{$\text{Disc}^{\text{Inst}}$}} & 55.36 & 46.78 & 39.67 & 26.67 & 42.12 \\
 & Semi-parametric w/ \texttt{\textbf{$\text{Disc}^{\text{FiD}}$}} & \textbf{56.50} & \textbf{49.23} & \textbf{42.46} & \textbf{28.42} & \textbf{44.15} \\
 \midrule
 \multirow{5}{*}{\shortstack{$k=4$}} 
 & Parametric (w/o Retrieval) & \multicolumn{4}{c}{36.50} & 36.50 \\
 & Semi-Parametric & 52.68 & 45.03 & 36.78 & 23.11 & 39.40 \\
 & Semi-Parametric w/ \texttt{\textbf{$\text{Disc}^{\text{Inst}}$}} & 55.03 & 46.34 & 39.95 & 27.54 & 42.21 \\
 & Semi-parametric w/ \texttt{\textbf{$\text{Disc}^{\text{FiD}}$}} & \textbf{56.45} & \textbf{49.18} & \textbf{41.86} & \textbf{29.07} & \textbf{44.14} \\
 \midrule
  \multirow{5}{*}{\shortstack{$k=5$}} 
 & Parametric (w/o Retrieval) & \multicolumn{4}{c}{36.83} & 36.83 \\
 & Semi-Parametric & 53.17 & 44.97 & 37.76 & 24.21 & 40.03 \\
 & Semi-Parametric w/ \texttt{\textbf{$\text{Disc}^{\text{Inst}}$}} & 54.19 & 45.63 & 38.41 & 26.78 & 41.26 \\
 & Semi-parametric w/ \texttt{\textbf{$\text{Disc}^{\text{FiD}}$}} & \textbf{56.28} & \textbf{49.18} & \textbf{41.64} & \textbf{28.63} & \textbf{43.93} \\
\bottomrule
\end{tabular}
\caption{GPT-3.5 results for ensembling over a different number of samples ($k$ is the number of in-context samples). Performance is reported in Exact Match (EM) on our entity-perturbed NQ-open \textbf{dev} set, according to the perturbation \% of retrieved documents.}
\label{table:ensemble_num_sample_results}
\end{table*}

In the Experiments (\S \ref{sec:eval_longpre} and Appendix \ref{sec:prompt_design}), we explain our use of ensemble strategy over the $k$ iterations and the marginalization over the top-10 generated answers to choose our final answer. One notable phenomenon spotted during our experiments is the ensemble's effect of improving over the simple average baseline (Table \ref{table:ensemble_result}). 

The ensemble strategy consistently outperforms the average setting across varying degrees of conflicting information. This suggests that not only does the ensemble of GPT-3.5 outputs alleviate the notorious sample variance issue, but it also enables the model to consider more probable output tokens across the iterations by avoiding the maximum likelihood outputs. One thing we would like to note is that our ensemble strategy demonstrates consistent patterns across various configurations (i.e., the number of samples) as shown in Table \ref{table:ensemble_num_sample_results}.

\subsection{Case Study on Perturbations}
\label{subsec:case_study}
In Table \ref{table:case_study} and Table \ref{table:case_study_ac}, we present side-by-side examples of documents of \textsc{MacNoise} against those of entity perturbation \cite{longpre-etal-2021-entity}. The comparison spans various perturbation types, namely Global Revision, Local Revision, Additional Context, and Entity Replacement w/ Context Match.

\paragraph{Global Revision.} We can see that \textsc{MacNoise} significantly restructures and updates the document's context to provide a more contemporary account. Specifically, it updates the narrative to reflect the events and performance of the Buffalo Bills during the 2020 season. This approach is comprehensive, ensuring the primary theme—how the Buffalo Bills performed during a particular season—remains consistent, but the details and timeline are considerably different. On the other hand, the Entity Replacement method, opts for a very specific and dramatic alteration. By replacing the year "1995" with "between 1652 and 1674," the document becomes factually incorrect but unnatural.

\paragraph{Local Revision.} Observations indicate that \textsc{MacNoise} entails nuanced changes tailored to fit an introduced narrative, while preserving the overarching theme. The founder of Victoria's Secret, originally "Roy Raymond," morphs into "John Thompson," with the surrounding context adjusted for coherence. While fundamental elements like the inception date and brand inspiration remain unchanged, specifics like names get modified. The Entity Replacement technique, in contrast, directly swaps "Roy Raymond" with "Patrick Denham," retaining the majority of the original narrative, which can result in potential mismatches. For example, the unchanged last name of Roy Raymond's wife might cause confusion.

\paragraph{Additional Context.} It becomes evident that the perturbation introduced by \textsc{MacNoise} provides an extended narrative, integrating not only changes in key entity details but also furnishing supplementary information that was not present in the original document. This seems to enrich the content, thereby providing more context, which makes it more realistic and challenging.
For instance, while the original narrative emphasizes Joe Spano's acting journey, the \textsc{MacNoise} perturbed version broadens the discourse, introducing Michael Thomas Grant's multifaceted talents and achievements. Conversely, the Entity Replacement strategy simply swaps "Joe Spano" for "Jaeden Lieberher," leaving the bulk of the content unaltered, which can lead to contextual mismatches.

\paragraph{Entity Replacement w/ Context Match.} \citet{longpre-etal-2021-entity}, directly replaces the name of the original entity, "Middle Island," with an unrelated entity, "Mid Glamorgan." This alteration results in a direct substitution without modifying the surrounding context, which can create inconsistencies. For instance, "Mid Glamorgan" does not correspond to any known location within Lake Erie, Ontario. Such discrepancies can make the document unrealistic, and not challenging to discern.
On the other hand, \textsc{MacNoise} also attempts an Entity Replacement but offers a slightly different name, "South Point Island," instead of "Middle Island." This change, while still altering the main entity, remains consistent with the rest of the context, providing a plausible and confounding alternative to the original entity.

\begin{table*}[h!]
\begin{center}
\begin{adjustbox}{width=1.0\textwidth}
\begin{tabular}{m{12cm}}
\toprule
\tiny{\texttt{\textit{\underline{Question}} \newline
when was the last time the bills won their division}} \\
\tiny{\texttt{\textit{\underline{Answer}} \newline
\textcolor{green}{1995}}} \\
\tiny{\texttt{\textit{\underline{Original Document}} \newline
title: 1995 Buffalo Bills season context: 1995 Buffalo Bills season The 1995 Buffalo Bills season was the 36th season for the club and its 26th in the National Football League. After suffering a losing season the previous season, the \textcolor{green}{1995} Bills won the AFC East, and made the playoffs for the seventh time in eight seasons. They defeated the Miami Dolphins in the Wild Card Game in Don Shula's final game as an NFL head coach. The following week they lost to the eventual AFC Champions, the Pittsburgh Steelers. The playoff win over Miami represents the last time to date that the Bills have won a
}} \\
\tiny{\texttt{\textit{\underline{Entity Replacement \cite{longpre-etal-2021-entity}}} \newline
title: \textcolor{red}{between 1652 and 1674} Buffalo Bills season context: \textcolor{red}{between 1652 and 1674} Buffalo Bills season The \textcolor{red}{between 1652 and 1674} Buffalo Bills season was the 36th season for the club and its 26th in the National Football League. After suffering a losing season the previous season, the \textcolor{red}{between 1652 and 1674} Bills won the AFC East, and made the playoffs for the seventh time in eight seasons. They defeated the Miami Dolphins in the Wild Card Game in Don Shula's final game as an NFL head coach. The following week they lost to the eventual AFC Champions, the Pittsburgh Steelers. The playoff win over Miami represents the last time to date that the Bills have won a
}} \\
\tiny{\texttt{\textit{\underline{MacNoise (Perturbation Type: Global Revision)}} \newline
title: 1995 Buffalo Bills season context: The 2020 Buffalo Bills season marked the 61st season for the club and its 51st in the National Football League Under the guidance of head coach Sean McDermott, the Bills improved on their 10-6 record from \textcolor{blue}{2019} and won their first AFC East division title since 1995, finishing with a 13-3 record. They made the playoffs for the seventh time in eight seasons and won their first playoff game since 1995, defeating the Indianapolis Colts in the Wild Card round. The Bills continued their playoff run by defeating the Baltimore Ravens in the Divisional round but ultimately lost to the Kansas City. Chiefs in the AFC Championship Game The 2020 season represents the last time to date that the Bills have won their division.
}} \\

\cmidrule(lr){1-1}

\tiny{\texttt{\textit{\underline{Question}} \newline
who was the creator of victoria's secret}} \\
\tiny{\texttt{\textit{\underline{Answer}} \newline
\textcolor{green}{Roy Raymond}}} \\
\tiny{\texttt{\textit{\underline{Original Document}} \newline
title: Victoria's Secret context: Victoria's Secret Victoria's Secret is an American designer, manufacturer, and marketer of women's lingerie, womenswear, and beauty products. Founded in 1977 as a response to packaged underwear, which the company's founder considered to be "ugly, floral-print nylon nightgowns", the company is now the largest American retailer of women's lingerie. Victoria's Secret was founded by \textcolor{green}{Roy Raymond}, and his wife Gaye Raymond, in San Francisco, California, on June 12, 1977. Eight years prior to founding Victoria's Secret, in the late 1960s, Raymond was embarrassed when purchasing lingerie for his wife at a department store. Newsweek reported him looking back on the
}} \\
\tiny{\texttt{\textit{\underline{Entity Replacement \cite{longpre-etal-2021-entity}}} \newline
title: Victoria's Secret context: Victoria's Secret Victoria's Secret is an American designer, manufacturer, and marketer of women's lingerie, womenswear, and beauty products. Founded in 1977 as a response to packaged underwear, which the company's founder considered to be "ugly, floral-print nylon nightgowns", the company is now the largest American retailer of women's lingerie. Victoria's Secret was founded by \textcolor{red}{Patrick Denham}, and his wife Gaye Raymond, in San Francisco, California, on June 12, 1977. Eight years prior to founding Victoria's Secret, in the late 1960s, Raymond was embarrassed when purchasing lingerie for his wife at a department store. Newsweek reported him looking back on the
}} \\
\tiny{\texttt{\textit{\underline{MacNoise (Perturbation Type: Local Revision)}} \newline
title: Victoria's Secret context: Victoria's Secret Victoria's Secret is an American designer, manufacturer, and marketer of women's lingerie, womenswear, and beauty products. The company was founded in 1977 by \textcolor{blue}{John Thompson} and his wife, Gaye \textcolor{blue}{Thompson}, in San Francisco, California. The idea for Victoria's Secret was born when John felt embarrassed while purchasing lingerie for his wife at a department store in the late 1960s. Recognizing the need for a more welcoming and specialized store for women's lingerie, the \textcolor{blue}{Thompsons} decided to create a store that would cater to this need. As a response to the "ugly, floral-print nylon nightgowns" that were commonly found in department stores, they designed Victoria's Secret to provide a comfortable and luxurious shopping experience for women, offering a wide range of lingerie, womenswear, and beauty products. Today, Victoria's Secret is the largest American retailer of women's lingerie.
}} \\

\bottomrule
\end{tabular}
\end{adjustbox}
\caption{Side-by-side example of the entity-perturbation \cite{longpre-etal-2021-entity} and \textsc{MacNoise} document samples. Text spans in \textcolor{green}{green} indicate the original answer text span, \textcolor{red}{red} is the entity-perturbed answer text span, including other text spans replaced by the perturbing sampled entity, and \textcolor{blue}{blue} is the LLM-perturbed answer-text span; we also mark other text spans in \textcolor{blue}{blue} in the \textsc{MacNoise} context if they are related to the perturbed answer.}
\label{table:case_study}
\end{center}
\end{table*}

\begin{table*}[h!]
\begin{center}
\begin{adjustbox}{width=1.0\textwidth}
\begin{tabular}{m{12cm}}
\toprule
\tiny{\texttt{\textit{\underline{Question}} \newline
who is the actor that plays tobias on nci}} \\
\tiny{\texttt{\textit{\underline{Answer}} \newline
\textcolor{green}{Joseph Peter Spano}}} \\
\tiny{\texttt{\textit{\underline{Original Document}} \newline
title: Joe Spano context: Joe Spano \textcolor{green}{Joseph Peter Spano} (born July 7, 1946) is an American actor who was originally best known for his role as Lt. Henry Goldblume on "Hill Street Blues". Spano is now best known for his work on "NCIS" as FBI Special Agent Tobias C. Fornell. Joe Spano was a member of the San Francisco improv group The Wing, and in college debuted as Paris in a production of "Romeo and Juliet" in 1967. In 1968, he helped found the Berkeley Repertory Theatre, appearing in its first production, and stayed with the company for 10 years. He moved to Hollywood
}} \\
\tiny{\texttt{\textit{\underline{Entity Replacement \cite{longpre-etal-2021-entity}}} 
\newline
title: Joe Spano context: Joe Spano \textcolor{red}{Jaeden Lieberher} (born July 7, 1946) is an American actor who was originally best known for his role as Lt. Henry Goldblume on "Hill Street Blues". Spano is now best known for his work on "NCIS" as FBI Special Agent Tobias C. Fornell. Joe Spano was a member of the San Francisco improv group The Wing, and in college debuted as Paris in a production of "Romeo and Juliet" in 1967. In 1968, he helped found the Berkeley Repertory Theatre, appearing in its first production, and stayed with the company for 10 years. He moved to Hollywood
}} \\
\tiny{\texttt{\textit{\underline{MacNoise (Perturbation Type: Additional Context)}} \newline
title: \textcolor{blue}{Michael Thomas Grant} context: \textcolor{blue}{Michael Thomas Grant}, born on July 7, 1946, is an American actor who gained initial recognition for his role as Lt Henry Goldblume on "Hill Street Blues". He is now best known for his portrayal of FBI Special Agent Tobias C Fornell on the popular television series "NCIS". Grant\'s acting journey began in San Francisco as a member of the improv group The Wing, and he made his college debut as Paris in a 1967 production of "Romeo and Juliet". In 1968, \textcolor{blue}{Grant} co-founded the Berkeley Repertory Theatre, where he appeared in its first production and remained with the company for a decade. Eventually, he moved to Hollywood and expanded his acting career to include television and film Some of his notable appearances include "Law \& Order: Special Victims Unit", "The Good Wife", and "Blue Bloods". Apart from acting, \textcolor{blue}{Grant} is a talented singer and musician, often displaying his abilities in stage productions and live performances.
}} \\

\cmidrule(lr){1-1}

\tiny{\texttt{\textit{\underline{Question}} \newline
what is the southern most part of canada}} \\
\tiny{\texttt{\textit{\underline{Answer}} \newline
\textcolor{green}{Middle Island}}} \\
\tiny{\texttt{\textit{\underline{Original Document}} \newline
title: Geography of Canada context: will likely cause significant changes to the environment, including loss of the polar bear, the exploration for resource then the extraction of these resources and an alternative transport route to the Panama Canal through the Northwest Passage. The northernmost point within the boundaries of Canada is Cape Columbia, Ellesmere Island, Nunavut . The northernmost point of the Canadian mainland is Zenith Point on Boothia Peninsula, Nunavut . The southernmost point is \textcolor{green}{Middle Island}, in Lake Erie, Ontario (41'41'N, 82'40'W); the southernmost water point lies just south of the island, on the Ontario–Ohio border (41'40'35"N). The southernmost point of the Canadian
}} \\
\tiny{\texttt{\textit{\underline{Entity Replacement \cite{longpre-etal-2021-entity}}} 
\newline
title: Geography of Canada context: will likely cause significant changes to the environment, including loss of the polar bear, the exploration for resource then the extraction of these resources and an alternative transport route to the Panama Canal through the Northwest Passage. The northernmost point within the boundaries of Canada is Cape Columbia, Ellesmere Island, Nunavut . The northernmost point of the Canadian mainland is Zenith Point on Boothia Peninsula, Nunavut . The southernmost point is \textcolor{red}{Mid Glamorgan}, in Lake Erie, Ontario (41'41'N, 82'40'W); the southernmost water point lies just south of the island, on the Ontario–Ohio border (41'40'35"N). The southernmost point of the Canadian
}} \\
\tiny{\texttt{\textit{\underline{MacNoise (Perturbation Type: Entity Replacement w/ Context Match)}} \newline
title: Geography of Canada context: The environment of Canada is undergoing significant changes, including the potential loss of the polar bear, increased exploration and extraction of resources, and the development of an alternative transport route to the Panama Canal through the Northwest Passage. The northernmost point within Canada's boundaries is Cape Columbia, Ellesmere Island, Nunavut, while the northernmost point of the Canadian mainland is Zenith Point on Boothia Peninsula, Nunavut. The southernmost point of Canada is \textcolor{blue}{South Point Island}, situated in Lake Erie, Ontario (41'41'N, 82'40'W). The southernmost water point lies just south of the island, along the Ontario-Ohio border (41'40'35"N).
}} \\

\bottomrule
\end{tabular}
\end{adjustbox}
\caption{Side-by-side example of the entity-perturbation \cite{longpre-etal-2021-entity} and \textsc{MacNoise} document samples. Text spans in \textcolor{green}{green} indicate the original answer text span, \textcolor{red}{red} is the entity-perturbed answer text span, including other text spans replaced by the perturbing sampled entity, and \textcolor{blue}{blue} is the LLM-perturbed answer-text span; we also mark other text spans in \textcolor{blue}{blue} in the \textsc{MacNoise} context if they are related to the perturbed answer.}
\label{table:case_study_ac}
\end{center}
\end{table*}

\end{document}